\documentclass[10pt,twocolumn,letterpaper]{article}

\usepackage{iccv}
\usepackage{times}
\usepackage{epsfig}
\usepackage{graphicx}
\usepackage{amsmath}
\usepackage{amssymb}
\usepackage{gensymb}
\usepackage{caption}
\usepackage{subcaption}
\usepackage{afterpage}
\usepackage{enumitem}
\usepackage[font=small]{caption}
\usepackage{hhline}
\usepackage{adjustbox}

\graphicspath{ {./figures/} {./figures/overlays/} {./figures/saliency/} {./figures/blur/} {./figures/map/} {./figures/7scenes/} }

\usepackage[breaklinks=true,bookmarks=false]{hyperref}

\iccvfinalcopy 

\def\iccvPaperID{981} 

\ificcvfinal\fi
\begin{document}

\title{PoseNet: A Convolutional Network for Real-Time 6-DOF Camera Relocalization}
\author{Alex Kendall \and Matthew Grimes \\
University of Cambridge\\
{\tt\small agk34, mkg30, rc10001 @cam.ac.uk}
\and Roberto Cipolla\\
}
\makeatletter
\let\@oldmaketitle\@maketitle
\renewcommand{\@maketitle}{\@oldmaketitle
\vspace{-3ex}
\makebox[\textwidth][c]{
   	\includegraphics[width=0.2\linewidth]{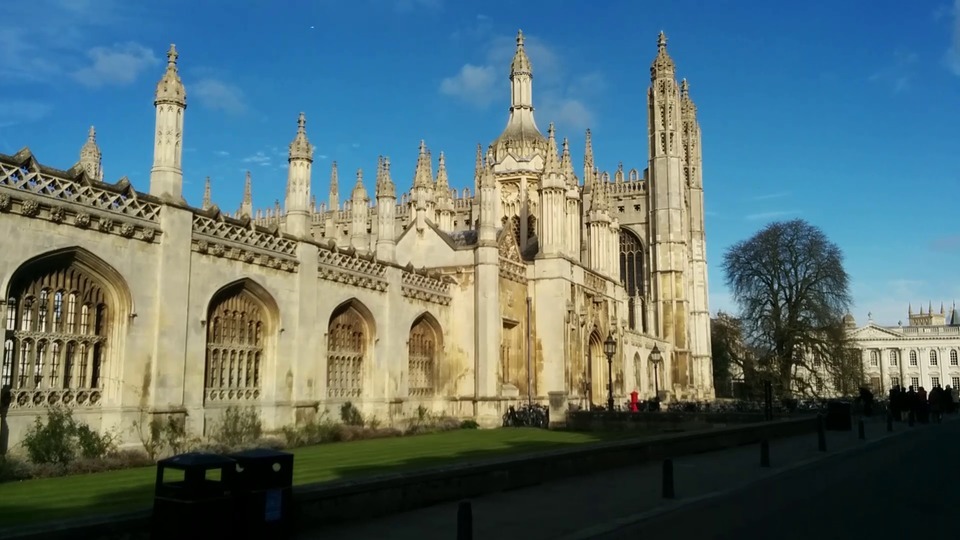}
   	\includegraphics[width=0.2\linewidth]{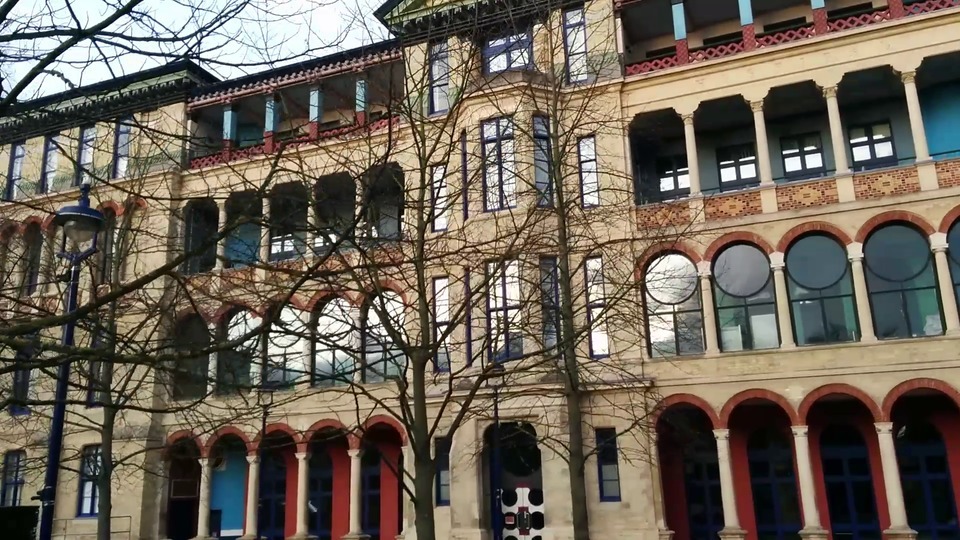}
   	\includegraphics[width=0.2\linewidth]{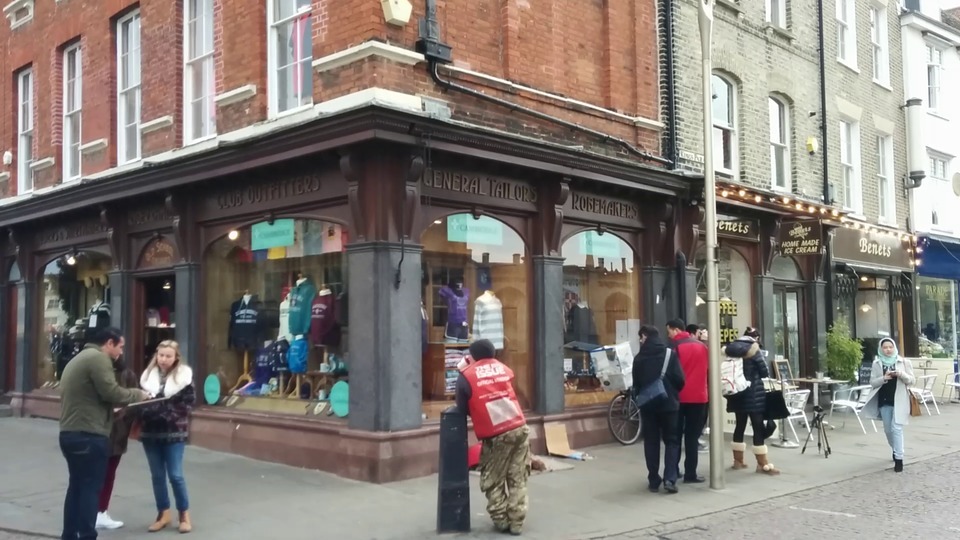}
   	\includegraphics[width=0.2\linewidth]{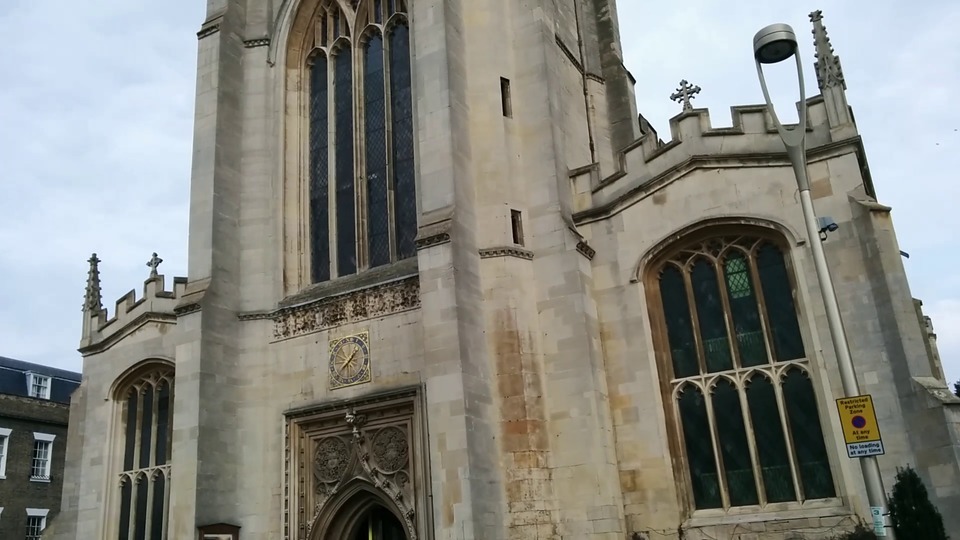}
   	}
\makebox[\textwidth][c]{
   	\includegraphics[width=0.2\linewidth]{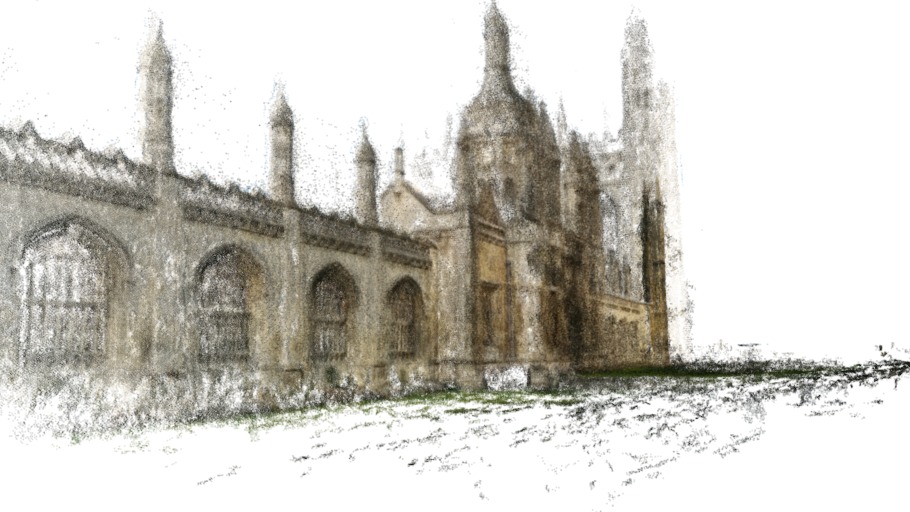}
   	\includegraphics[width=0.2\linewidth]{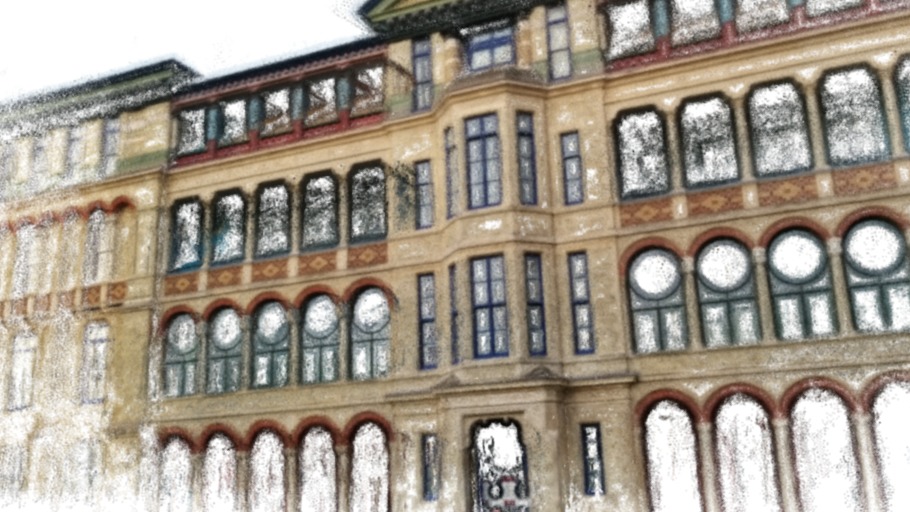}
   	\includegraphics[width=0.2\linewidth]{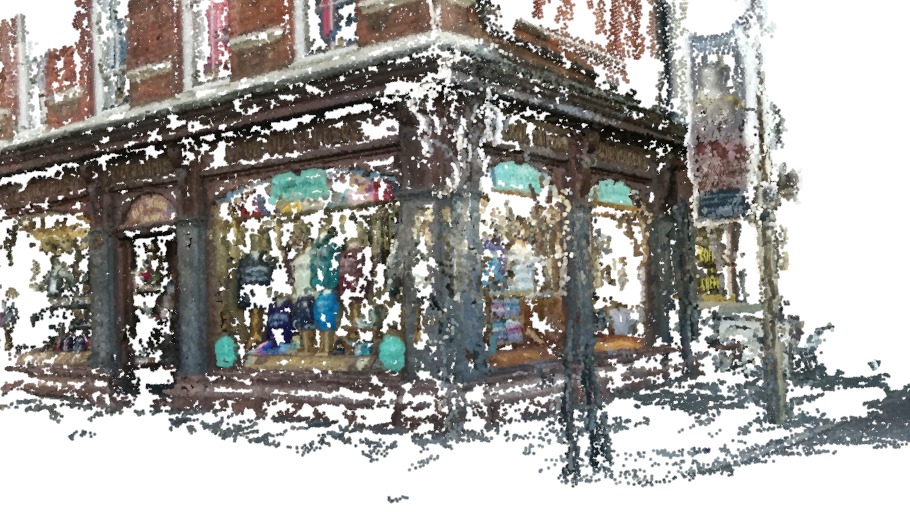}
   	\includegraphics[width=0.2\linewidth]{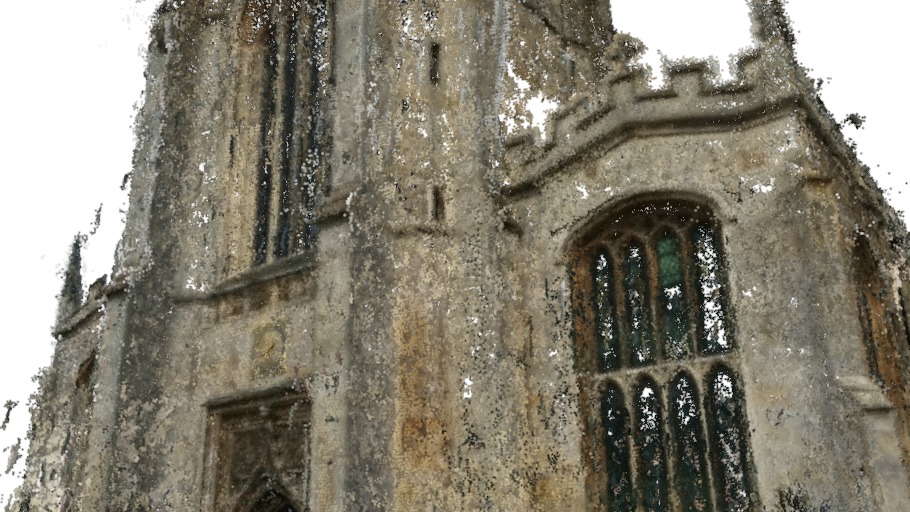}
   	}
\makebox[\textwidth][c]{
   	\includegraphics[width=0.2\linewidth]{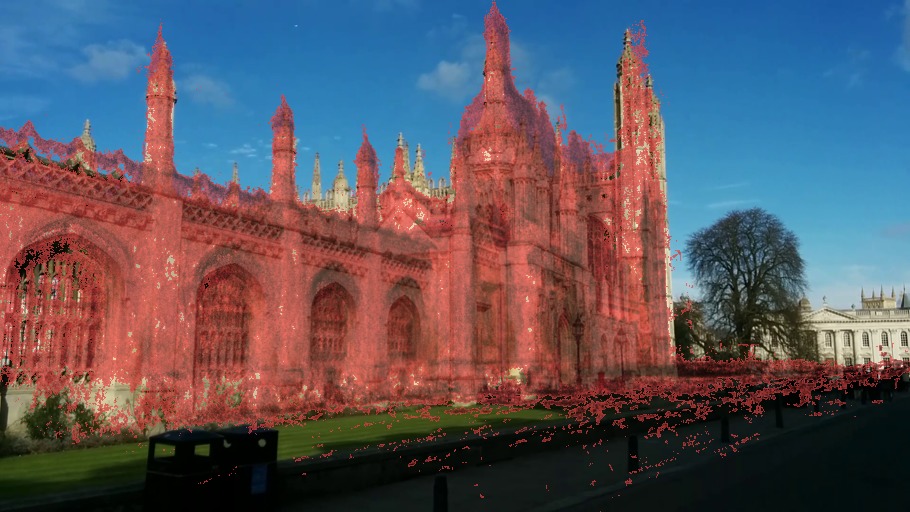}
   	\includegraphics[width=0.2\linewidth]{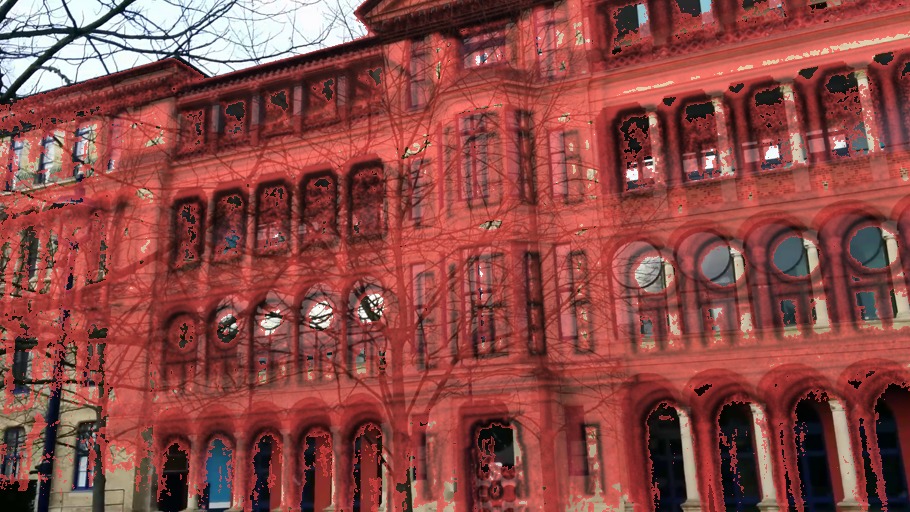}
   	\includegraphics[width=0.2\linewidth]{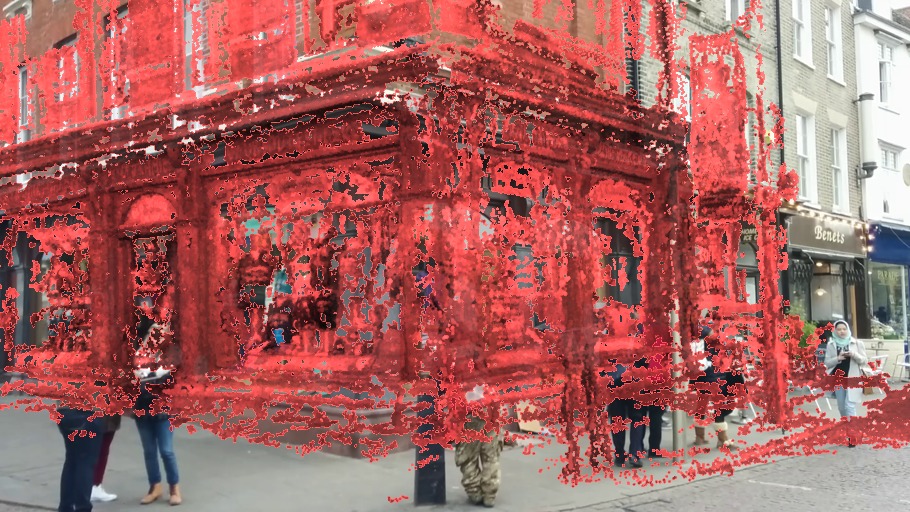}
   	\includegraphics[width=0.2\linewidth]{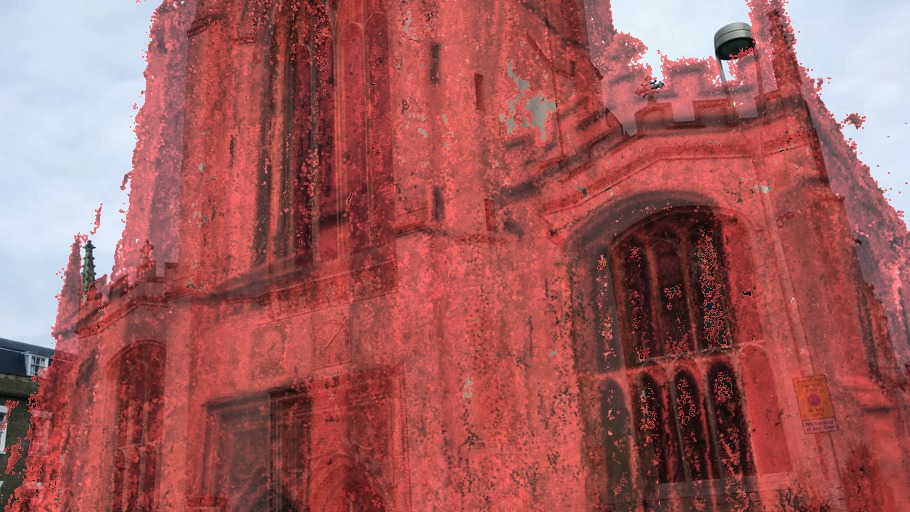}
   	}
\makebox[\textwidth][c]{
\makebox[0.2\textwidth][c]{
\textit{King's College}
   	}
\makebox[0.2\textwidth][c]{
\textit{Old Hospital}
   	}
\makebox[0.2\textwidth][c]{
\textit{Shop Fa\c cade}
   	}
\makebox[0.2\textwidth][c]{
\textit{St Mary's Church}
   	}}\\
 \refstepcounter{figure}\normalfont Figure~\thefigure: \textbf{PoseNet: Convolutional neural network monocular camera relocalization.} Relocalization results for an input image (top), the predicted camera pose of a visual reconstruction (middle), shown again overlaid in red on the original image (bottom). Our system relocalizes to within approximately $2m$ and $6\degree$ for large outdoor scenes spanning $50,000m^2$. For an online demonstration, please see our project webpage:~ {\tt\small\href{http://mi.eng.cam.ac.uk/projects/relocalisation/}{mi.eng.cam.ac.uk/projects/relocalisation/}}
\\
\label{fig:teaser}}
\makeatother
\maketitle
\thispagestyle{empty}

\begin{abstract}
We present a robust and real-time monocular six degree of freedom relocalization system. Our system trains a convolutional neural network to regress the 6-DOF camera pose from a single RGB image in an end-to-end manner with no need of additional engineering or graph optimisation. The algorithm can operate indoors and outdoors in real time, taking 5ms per frame to compute. It obtains approximately 2m and 6\degree accuracy for large scale outdoor scenes and 0.5m and 10\degree accuracy indoors. This is achieved using an efficient 23 layer deep convnet, demonstrating that convnets can be used to solve complicated out of image plane regression problems. This was made possible by leveraging transfer learning from large scale classification data. We show that the PoseNet localizes from high level features and is robust to difficult lighting, motion blur and different camera intrinsics where point based SIFT registration fails. Furthermore we show how the pose feature that is produced generalizes to other scenes allowing us to regress pose with only a few dozen training examples.
\end{abstract}

\section{Introduction}

Inferring where you are, or localization, is crucial for mobile robotics, navigation and augmented reality. This paper addresses the lost or kidnapped robot problem by introducing a novel relocalization algorithm. Our proposed system, PoseNet, takes a single 224x224 RGB image and regresses the camera's 6-DoF pose relative to a scene. Fig.~\ref{fig:teaser} demonstrates some examples. The algorithm is simple in the fact that it consists of a convolutional neural network (convnet) trained end-to-end to regress the camera's orientation and position. It operates in real time, taking 5ms to run, and obtains approximately 2m and 6 degrees accuracy for large scale outdoor scenes (covering a ground area of up to $50,000m^2$). 

Our main contribution is the deep convolutional neural network camera pose regressor. We introduce two novel techniques to achieve this. We leverage transfer learning from recognition to relocalization with very large scale classification datasets. Additionally we use structure from motion to automatically generate training labels (camera poses) from a video of the scene. This reduces the human labor in creating labeled video datasets to just recording the video. 

Our second main contribution is towards understanding the representations that this convnet generates. We show that the system learns to compute feature vectors which are easily mapped to pose, and which also generalize to unseen scenes with a few additional training samples.

Appearance-based relocalization has had success \cite{cummins2008fab,sunderhauf2015performance} in coarsely locating the camera among a limited, discretized set of place labels, leaving the pose estimation to a separate system. This paper presents a means of computing continuous pose directly from appearance. The scene may include multiple objects and need not be viewed under consistent conditions. For example the scene may include dynamic objects like people and cars or experience changing weather conditions. 

Simultaneous localization and mapping (SLAM) is a traditional solution to this problem. We introduce a new framework for localization which removes several issues faced by typical SLAM pipelines, such as the need to store densely spaced keyframes, the need to maintain separate mechanisms for appearance-based localization and landmark-based pose estimation, and a need to establish frame-to-frame feature correspondence. We do this by mapping monocular images to a high-dimensional representation that is robust to nuisance variables. We empirically show that this representation is a smoothly varying injective (one-to-one) function of pose, allowing us to regress pose directly from the image without need of tracking.

Training convolutional networks is usually dependent on very large labeled image datasets, which are costly to assemble. Examples include the \textit{ImageNet} \cite{deng2009imagenet} and \textit{Places} \cite{zhou2014learning} datasets, with 14 million and 7 million hand-labeled images, respectively. We employ two techniques to overcome this limitation:
\begin{itemize}[noitemsep]
\item an automated method of labeling data using structure from motion to generate large regression datasets of camera pose 
\item transfer learning which trains a pose regressor, pre-trained as a classifier, on immense image recognition datasets. This converges to a lower error in less time, even with a very sparse training set, as compared to training from scratch.
\end{itemize}

\section{Related work}

There are generally two approaches to localization: metric and appearance-based. 
Metric SLAM localizes a mobile robot by focusing on creating a sparse \cite{klein2007parallel,kaess2011isam2} or dense \cite{newcombe2011dtam,engel2014lsd} map of the environment. Metric SLAM estimates the camera's continuous pose, given a good initial pose estimate. Appearance-based localization provides this coarse estimate by classifying the scene among a limited number of discrete locations. Scalable appearance-based localizers have been proposed such as \cite{cummins2008fab} which uses SIFT features \cite{lowe2004distinctive} in a bag of words approach to probabilistically recognize previously viewed scenery. Convnets have also been used to classify a scene into one of several location labels \cite{sunderhauf2015performance}. Our approach combines the strengths of these approaches: it does not need an initial pose estimate, and produces a continuous pose. Note we do not build a map, rather we train a neural network, whose size, unlike a map, does not require memory linearly proportional to the size of the scene (see fig.~\ref{fig:speed}).

Our work most closely follows from the Scene Coordinate Regression Forests for relocalization proposed in \cite{shotton2013scene}. This algorithm uses depth images to create scene coordinate labels which map each pixel from camera coordinates to global scene coordinates. This was then used to train a regression forest to regress these labels and localize the camera. However, unlike our approach, this algorithm is limited to RGB-D images to generate the scene coordinate label, in practice constraining its use to indoor scenes.

Previous research such as \cite{wang2006coarse,li2012worldwide,hao20123d,bergamo2013leveraging} has also used SIFT-like point based features to match and localize from landmarks. However these methods require a large database of features and efficient retrieval methods. A method which uses these point features is structure from motion (SfM) \cite{wu2013towards,agarwal2011building,snavely2006photo} which we use here as an offline tool to automatically label video frames with camera pose. We use \cite{furukawa2010towards} to generate a dense visualisation of our relocalization results.

Despite their ability in classifying spatio-temporal data, convolutional neural networks are only just beginning to be used for regression. They have advanced the state of the art in object detection \cite{szegedy2014going} and human pose regression \cite{toshev2014deeppose}. However these have limited their regression targets to lie in the 2-D image plane. Here we demonstrate regressing the full 6-DOF camera pose transform including depth and out-of-plane rotation. Furthermore, we show we are able to learn regression as opposed to being a very fine resolution classifier.

It has been shown that convnet representations trained on classification problems generalize well to other tasks \cite{razavian2014cnn,oquab2014learning,bengio2013representation,donahue2013decaf}. We show that you can apply these representations of classification to 6-DOF regression problems. Using these pre-learned representations allows convnets to be used on smaller datasets without overfitting.

\section{Model for deep regression of camera pose}
In this section we describe the convolutional neural network (convnet) we train to estimate camera pose directly from a monocular image, $I$. Our network outputs a pose vector $\mathbf{p}$, given by a 3D camera position $\mathbf{x}$ and orientation represented by quaternion $\mathbf{q}$:
\begin{equation}
\mathbf{p} = [\mathbf{x}, \mathbf{q}]
\end{equation}
Pose $\mathbf{p}$ is defined relative to an arbitrary global reference frame. We chose quaternions as our orientation representation, because arbitrary 4-D values are easily mapped to legitimate rotations by normalizing them to unit length. This is a simpler process than the orthonormalization required of rotation matrices.

\subsection{Simultaneously learning location and\\orientation}
To regress pose, we train the convnet on Euclidean loss using stochastic gradient descent with the following objective loss function:
\begin{equation}
loss(I) = \left\lVert\mathbf{\hat{x}} - \mathbf{x}\right\rVert_2 + \beta \left\lVert \mathbf{\hat{q}}-\frac{\mathbf{q}}{\left\lVert\mathbf{q}\right\rVert}\right\rVert_2
\label{eqn:loss}
\end{equation}
Where $\beta$ is a scale factor chosen to keep the expected value of position and orientation errors to be approximately equal. 

The set of rotations lives on the unit sphere in quaternion space. However the Euclidean loss function makes no effort to keep $\mathbf{q}$ on the unit sphere. We find, however, that during training, $\mathbf{q}$ becomes close enough to $\mathbf{\hat{q}}$ such that the distinction between spherical distance and Euclidean distance becomes insignificant. For simplicity, and to avoid hampering the optimization with unnecessary constraints, we chose to omit the spherical constraint.
 
\begin{figure}[t]
\begin{center}
   	\includegraphics[width=0.7\linewidth]{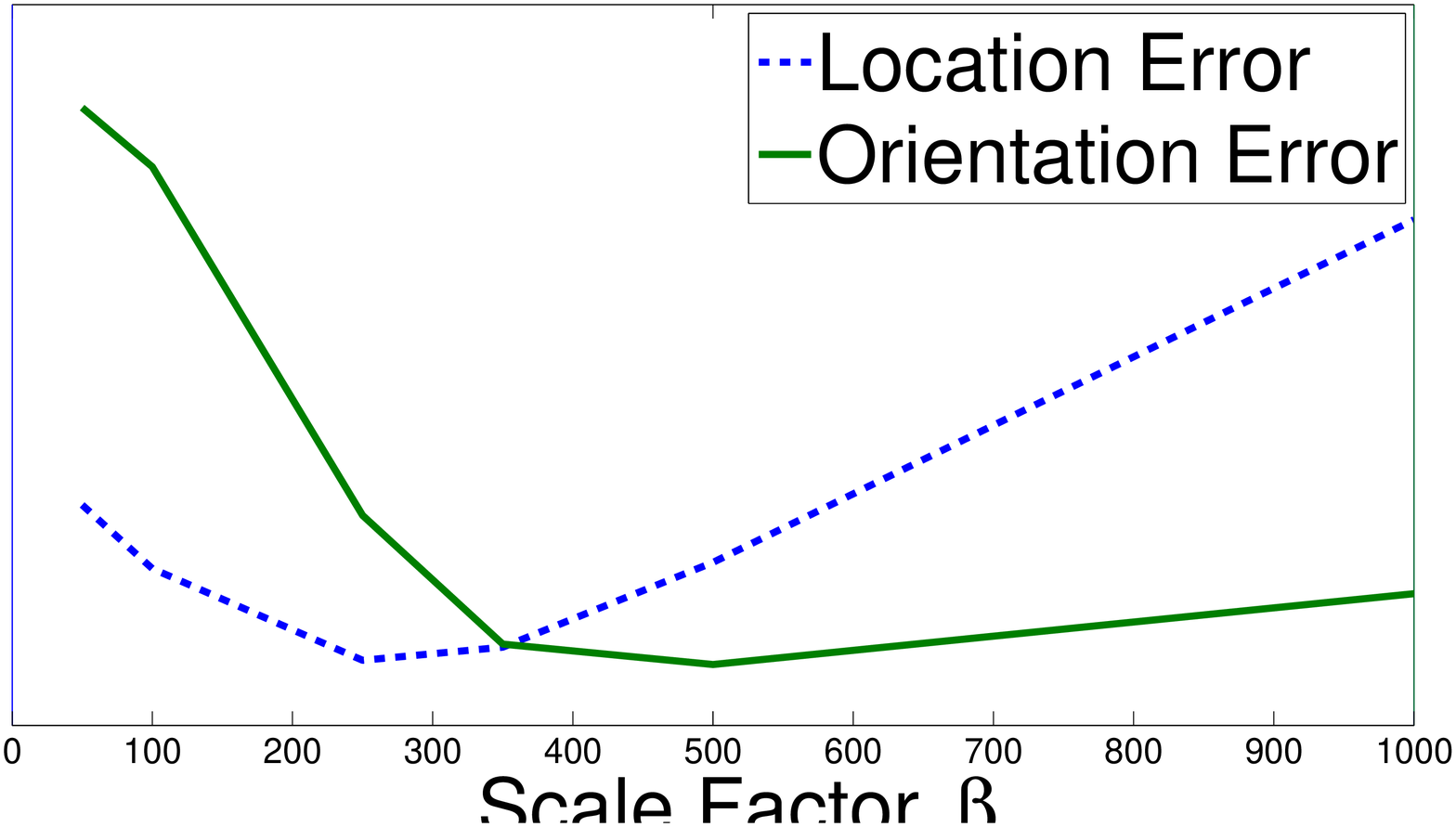}
\end{center}
   \caption{Relative performance of position and orientation regression on \textbf{a single convnet with a range of scale factors} for an indoor scene, Chess. This demonstrates that learning with the optimum scale factor leads to the convnet uncovering a more accurate pose function.}
\label{fig:scalefactor}
\end{figure}

We found that training individual networks to regress position and orientation separately performed poorly compared to when they were trained with full 6-DOF pose labels (fig.~\ref{fig:scalefactor}). With just position, or just orientation information, the convnet was not as effectively able to determine the function representing camera pose. We also experimented with branching the network lower down into two separate components to regress position and orientation. However, we found that it too was less effective, for similar reasons: separating into distinct position and orientation regressors denies each the information necessary to factor out orientation from position, or vice versa.

In our loss function (\ref{eqn:loss}) a balance $\beta$ must be struck between the orientation and translation penalties (fig.~\ref{fig:scalefactor}). They are highly coupled as they are regressed from the same model weights. We observed that the optimal $\beta$ was given by the ratio between expected error of position and orientation at the end of training, not the beginning. We found $\beta$ to be greater for outdoor scenes as position errors tended to be relatively greater. Following this intuition we fine tuned $\beta$ using grid search. For the indoor scenes it was between 120 to 750 and outdoor scenes between 250 to 2000.

We found it was important to randomly initialize the final position regressor layer so that the norm of the weights corresponding to each position dimension was proportional to that dimension's spatial extent.

Classification problems have a training example for every category. This is not possible for regression as the output is continuous and infinite. Furthermore, other convnets that have been used for regression operate off very large datasets \cite{toshev2014deeppose,sermanet2013overfeat}. For localization regression to work off limited data we leverage the powerful representations learned off these large classification datasets by pretraining the weights on these datasets.

\subsection{Architecture}

For the experiments in this paper we use a state of the art deep neural network architecture for classification, GoogLeNet \cite{szegedy2014going}, as a basis for developing our pose regression network. GoogLeNet is a 22 layer convolutional network with six `inception modules' and two additional intermediate classifiers which are discarded at test time. Our model is a slightly modified version of GoogLeNet with 23 layers (counting only the layers with trainable parameters). We modified GoogLeNet as follows:
\begin{itemize}[noitemsep]
\item Replace all three softmax classifiers with affine regressors. The softmax layers were removed and each final fully connected layer was modified to output a pose vector of 7-dimensions representing position (3) and orientation (4). 
\item Insert another fully connected layer before the final regressor of feature size 2048. This was to form a localization feature vector which may then be explored for generalisation. 
\item At test time we also normalize the quaternion orientation vector to unit length.
\end{itemize}
We rescaled the input image so that the smallest dimension was 256 pixels before cropping to the 224x224 pixel input to the GoogLeNet convnet. The convnet was trained on random crops (which do not affect the camera pose). At test time we evaluate it with both a single center crop and also densely with 128 uniformly spaced crops of the input image, averaging the resulting pose vectors. With parallel GPU processing, this results in a computational time increase from 5ms to 95ms per image.

We experimented with rescaling the original image to different sizes before cropping for training and testing. Scaling up the input is equivalent to cropping the input before downsampling to 256 pixels on one side. This increases the spatial resolution of the input pixels. We found that this does not increase the localization performance, indicating that context and field of view is more important than resolution for relocalization.

The PoseNet model was implemented using the Caffe library \cite{jia2014caffe}. It was trained using stochastic gradient descent with a base learning rate of $10^-5$, reduced by 90\% every 80 epochs and with momentum of 0.9. Using one half of a dual-GPU card (NVidia Titan Black), training took an hour using a batch size of 75. For reasons of time, we did not explore multi-GPU training, although it is reasonable to expect better results from using double the throughput and memory. We subtracted a separate image mean for each scene as we found this to improve experimental performance. 

\section{Dataset}

\begin{figure}[t]
   	\includegraphics[width=\linewidth]{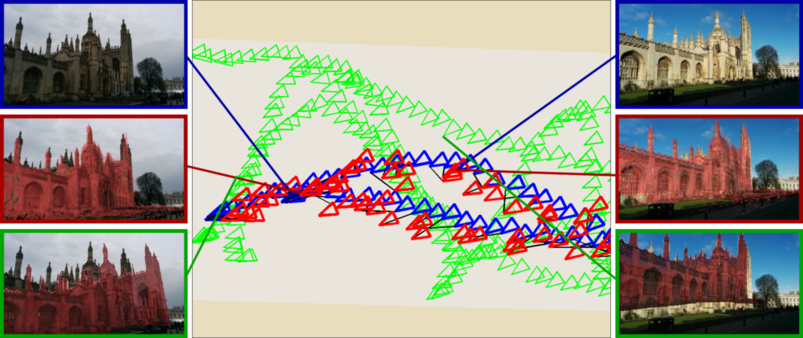}
   \caption{Magnified view of a sequence of \textbf{training (green)} and \textbf{testing (blue)} cameras for King's College. We show the \textbf{predicted camera pose in red} for each testing frame. The images show the test image (top), the predicted view from our convnet overlaid in red on the input image (middle) and the nearest neighbour training image overlaid in red on the input image (bottom). This shows our system can interpolate camera pose effectively in space between training frames.}
\label{fig:zoommap}
\end{figure}

\begin{figure*}[t]
\makebox[\textwidth][c]{
   	\includegraphics[height=0.2\linewidth]{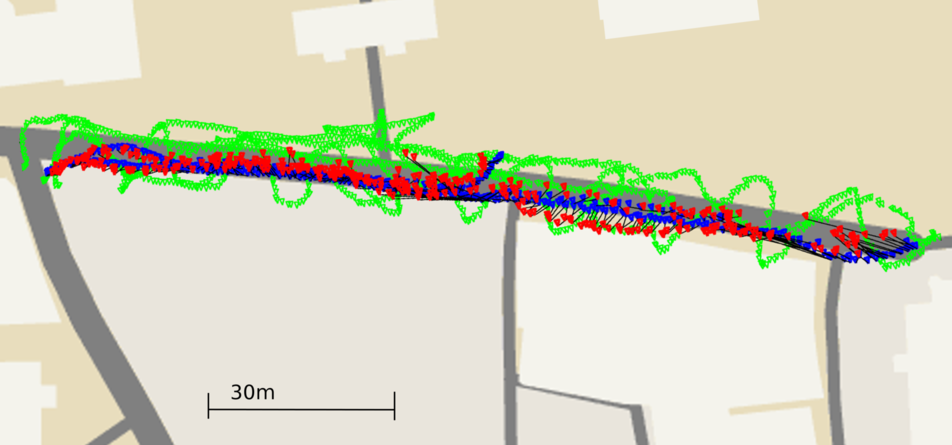}
   	~
   	\includegraphics[height=0.2\linewidth]{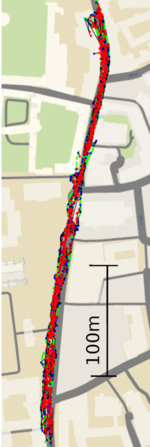}
   	~
   	\includegraphics[height=0.2\linewidth]{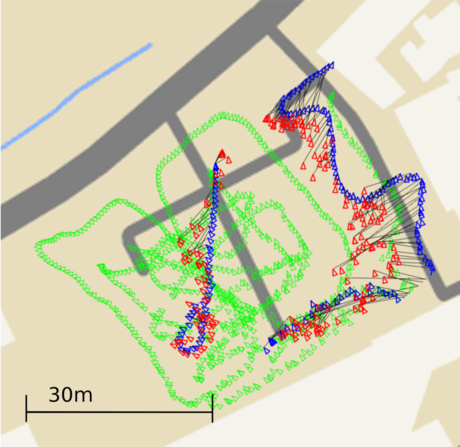}
   	~
   	\includegraphics[height=0.2\linewidth]{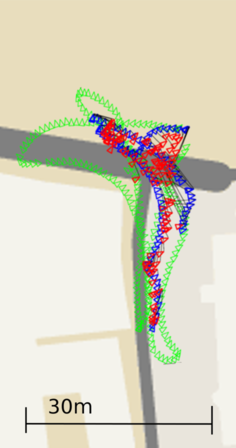}
   	~
   	\includegraphics[height=0.2\linewidth]{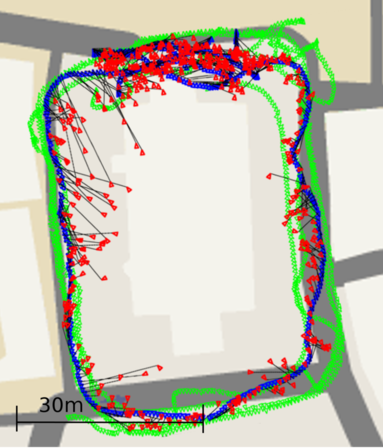}
   	}
\makebox[\textwidth][l]{
\makebox[0.44\textwidth][c]{
\textit{King's College}
   	}
\makebox[0.03\textwidth][c]{
\textit{Street}
   	}
\makebox[0.24\textwidth][c]{
\textit{Old Hospital}
   	}
\makebox[0.1\textwidth][c]{
\textit{Shop Fa\c cade}
   	}
\makebox[0.2\textwidth][c]{
\textit{St Mary's Church}
   	}}
   \caption{\textbf{Map of dataset} showing training frames (green), testing frames (blue) and their predicted camera pose (red). The testing sequences are distinct trajectories from the training sequences and each scene covers a very large spatial extent.}
   \label{fig:map}
   
\makebox[\textwidth][c]{
   	\includegraphics[width=0.12\linewidth]{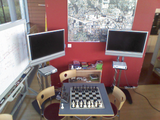}
   	\includegraphics[width=0.12\linewidth]{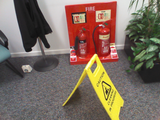}
   	\includegraphics[width=0.12\linewidth]{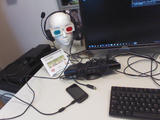}
   	\includegraphics[width=0.12\linewidth]{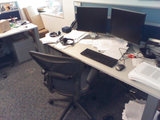}
   	\includegraphics[width=0.12\linewidth]{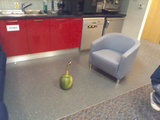}
   	\includegraphics[width=0.12\linewidth]{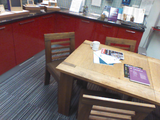}
   	\includegraphics[width=0.12\linewidth]{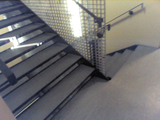}
   	}
   \caption{\textbf{7 Scenes dataset} example images from left to right; Chess, Fire, Heads, Office, Pumpkin, Red Kitchen and Stairs.}
   \label{fig:7scenes}
\begin{tabular}{l|c c|c|c c c c}
 & \multicolumn{2}{|c|}{\# Frames} & Spatial & SCoRe Forest & Dist. to Conv. & & \\
Scene & Train & Test & Extent (m) & (Uses RGB-D) & Nearest Neighbour & PoseNet & Dense PoseNet\\
\hhline{=|==|=|====}
King's College 	& 1220 & 343 & 140 x 40m 	& N/A 				& 3.34m, 5.92\degree & 1.92m, 5.40\degree & 1.66m, 4.86\degree \\
Street & 3015 	& 2923 & 500 x 100m 			& N/A 				& 1.95m, 9.02\degree & 3.67m, 6.50\degree & 2.96m, 6.00\degree \\
Old Hospital 	& 895 & 182 & 50 x 40m 		& N/A 				& 5.38m, 9.02\degree & 2.31m, 5.38\degree & 2.62m, 4.90\degree \\
Shop Fa\c cade 	& 231 & 103 & 35 x 25m 		& N/A 				& 2.10m, 10.4\degree & 1.46m, 8.08\degree & 1.41m, 7.18\degree\\
St Mary's Church & 1487 & 530 & 80 x 60m 	& N/A 				& 4.48m, 11.3\degree & 2.65m, 8.48\degree & 2.45m, 7.96\degree\\
\hline
Chess 		& 4000 & 2000 & 3 x 2 x 1m 		& 0.03m, 0.66\degree & 0.41m, 11.2\degree & 0.32m, 8.12\degree & 0.32m, 6.60\degree\\
Fire 		& 2000 & 2000 & 2.5 x 1 x 1m 	& 0.05m, 1.50\degree & 0.54m, 15.5\degree & 0.47m, 14.4\degree & 0.47m, 14.0\degree \\
Heads 		& 1000 & 1000 & 2 x 0.5 x 1m 	& 0.06m, 5.50\degree & 0.28m, 14.0\degree & 0.29m, 12.0\degree & 0.30m, 12.2\degree \\
Office 		& 6000 & 4000 & 2.5 x 2 x 1.5m 	& 0.04m, 0.78\degree & 0.49m, 12.0\degree & 0.48m, 7.68\degree & 0.48m, 7.24\degree \\
Pumpkin 		& 4000 & 2000 & 2.5 x 2 x 1m 	& 0.04m, 0.68\degree & 0.58m, 12.1\degree & 0.47m, 8.42\degree & 0.49m, 8.12\degree \\
Red Kitchen 	& 7000 & 5000 & 4 x 3 x 1.5m 	& 0.04m, 0.76\degree & 0.58m, 11.3\degree & 0.59m, 8.64\degree & 0.58m, 8.34\degree \\
Stairs 		& 2000 & 1000 & 2.5 x 2 x 1.5m 	& 0.32m, 1.32\degree & 0.56m, 15.4\degree & 0.47m, 13.8\degree & 0.48m, 13.1\degree \\
\end{tabular}
\caption{\textbf{Dataset details and results.} We show median performance for PoseNet on all scenes, evaluated on a single 224x224 center crop and 128 uniformly separated dense crops.
For comparison we plot the results from SCoRe Forest \cite{shotton2013scene} which uses depth, therefore fails on outdoor scenes. This system regresses pixel-wise world coordinates of the input image at much larger resolution. This requires a dense depth map for training and an extra RANSAC step to determine the camera's pose. Additionally, we compare to matching the nearest neighbour feature vector representation from PoseNet. This demonstrates our regression PoseNet performs better than a classifier. }
   \label{fig:mainresults}
\makebox[\textwidth][c]{
\begin{subfigure}[b]{0.26\linewidth}
   	\adjustbox{trim={0.09\width} {0} {0.08\width} {0},clip}{\includegraphics[width=1.2\linewidth]{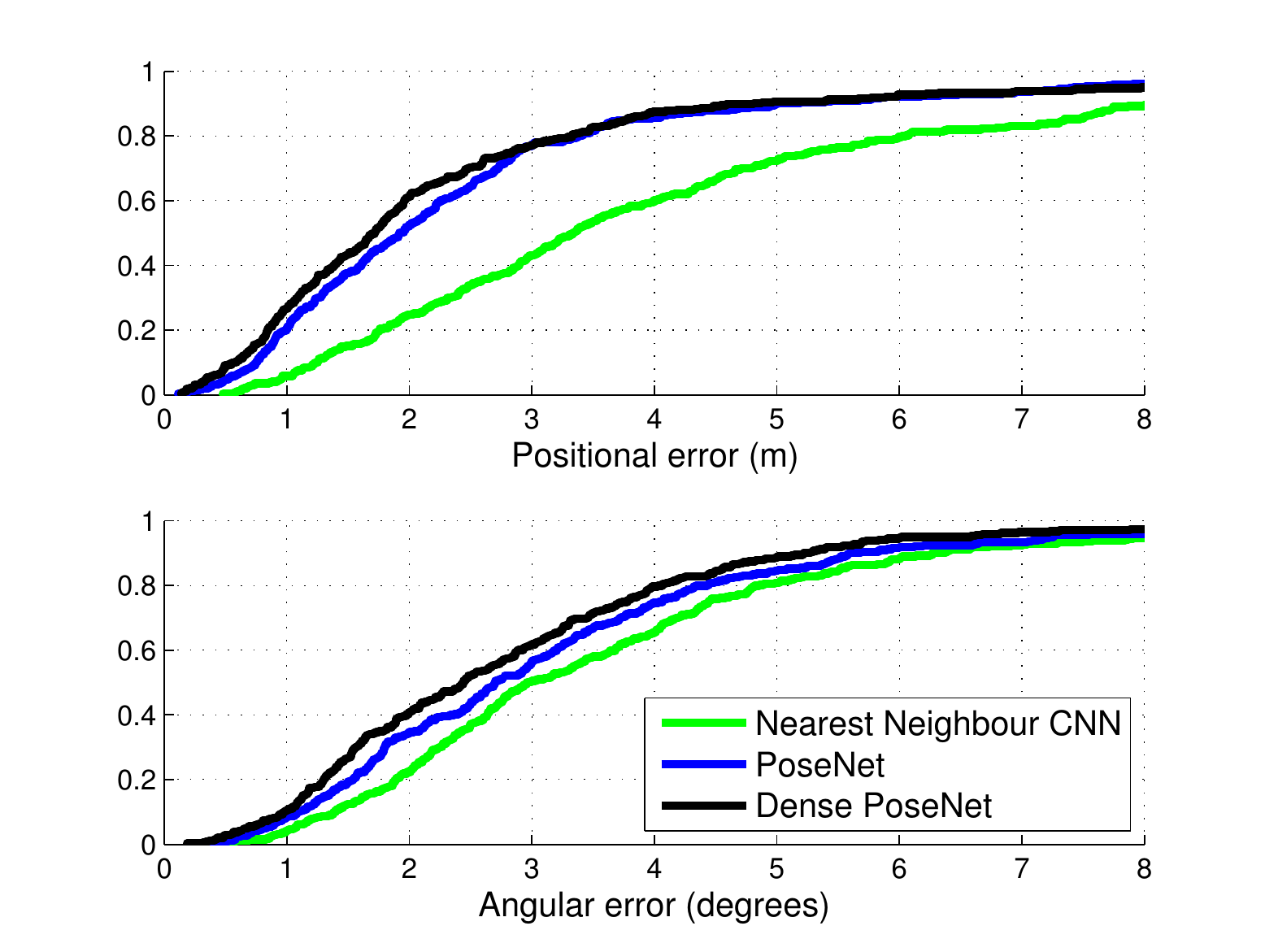}}
   	\caption{King's College}
\end{subfigure}
\begin{subfigure}[b]{0.26\linewidth}
   	\adjustbox{trim={0.09\width} {0} {0.08\width} {0},clip}{\includegraphics[width=1.2\linewidth]{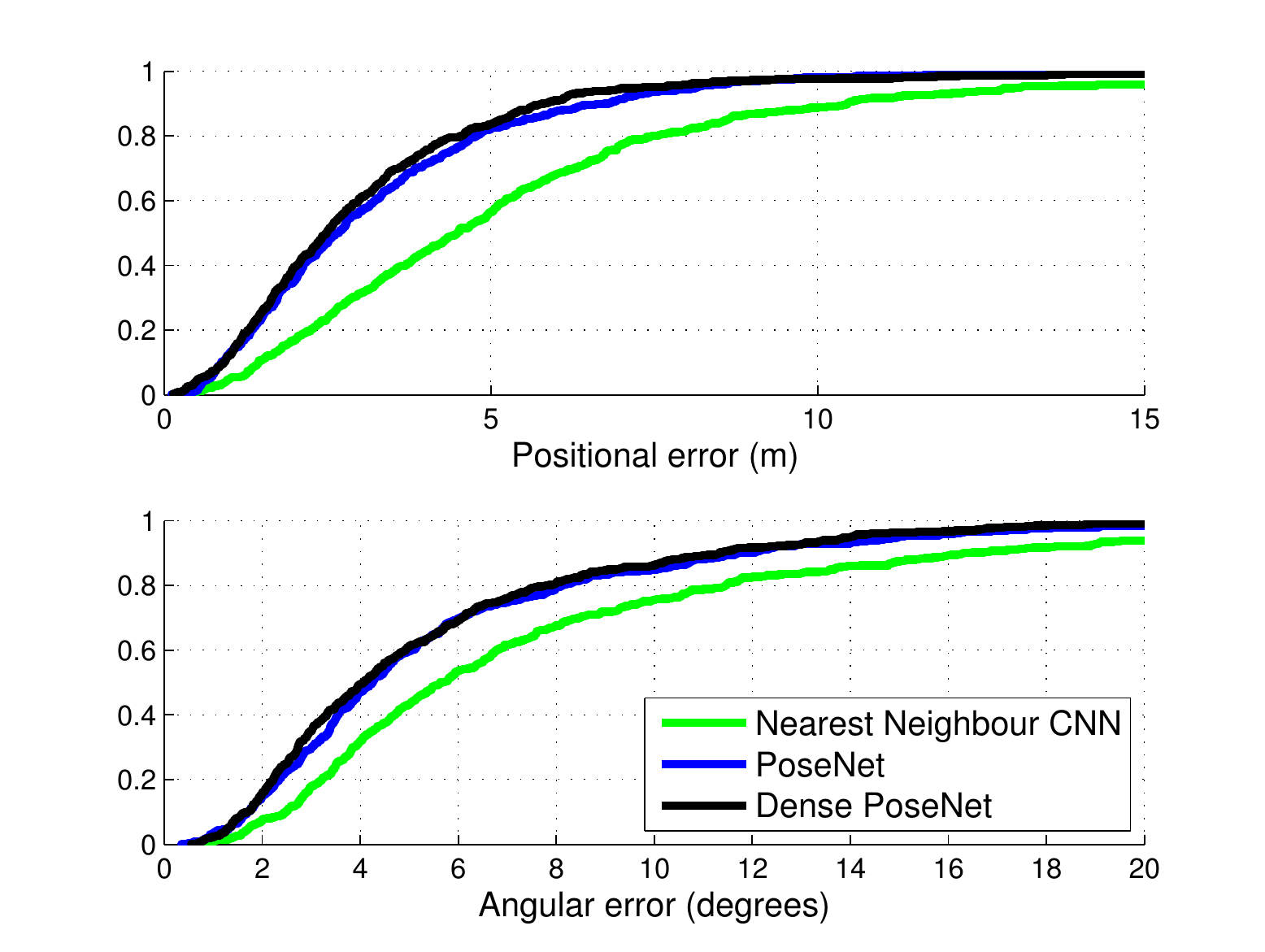}}
   	\caption{St Mary's Church}
\end{subfigure}
\begin{subfigure}[b]{0.26\linewidth}
   	\adjustbox{trim={0.09\width} {0} {0.08\width} {0},clip}{\includegraphics[width=1.2\linewidth]{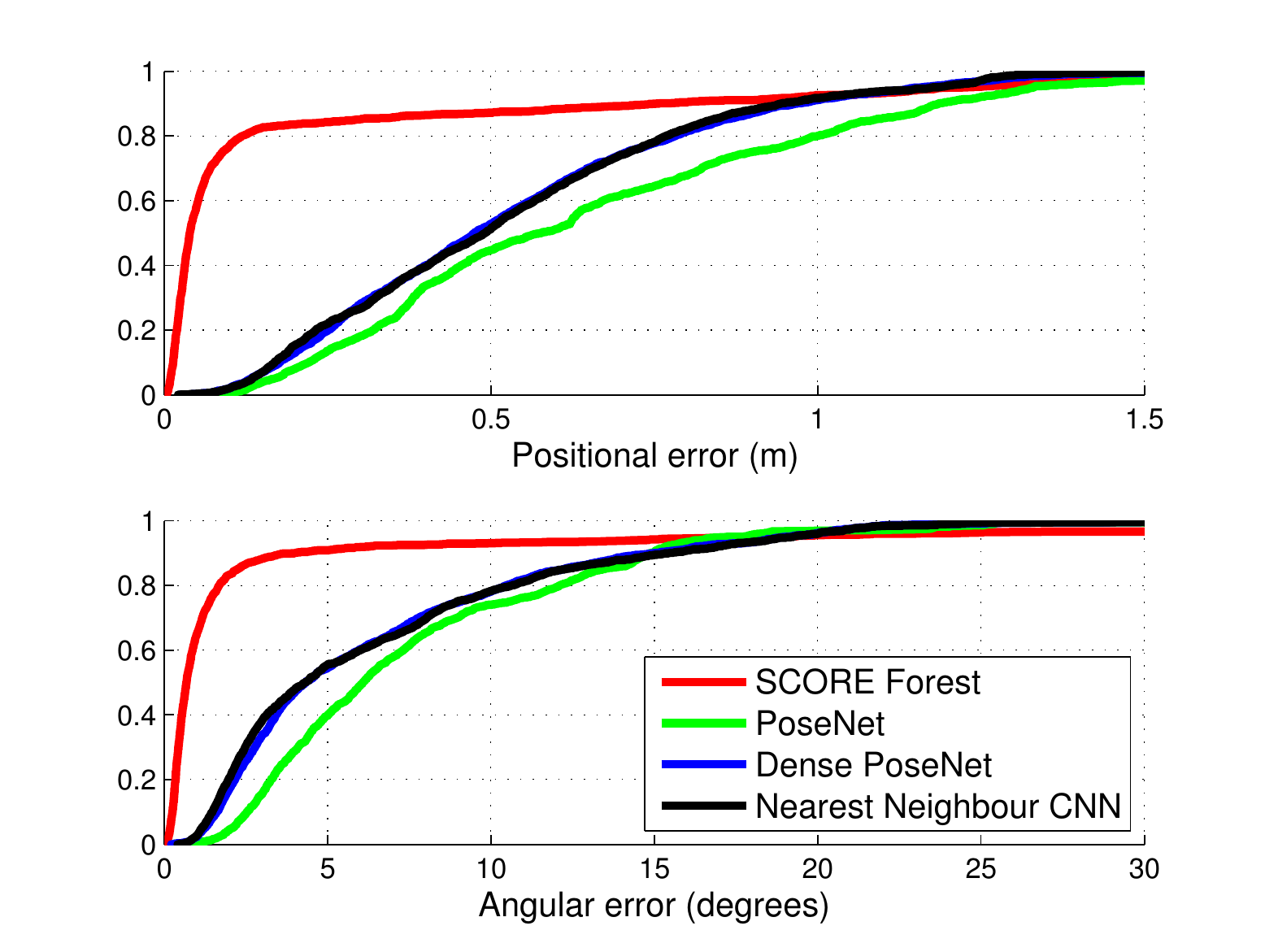}}
   	\caption{Pumpkin}
\end{subfigure}
\begin{subfigure}[b]{0.26\linewidth}
   	\adjustbox{trim={0.09\width} {0} {0.08\width} {0},clip}{\includegraphics[width=1.2\linewidth]{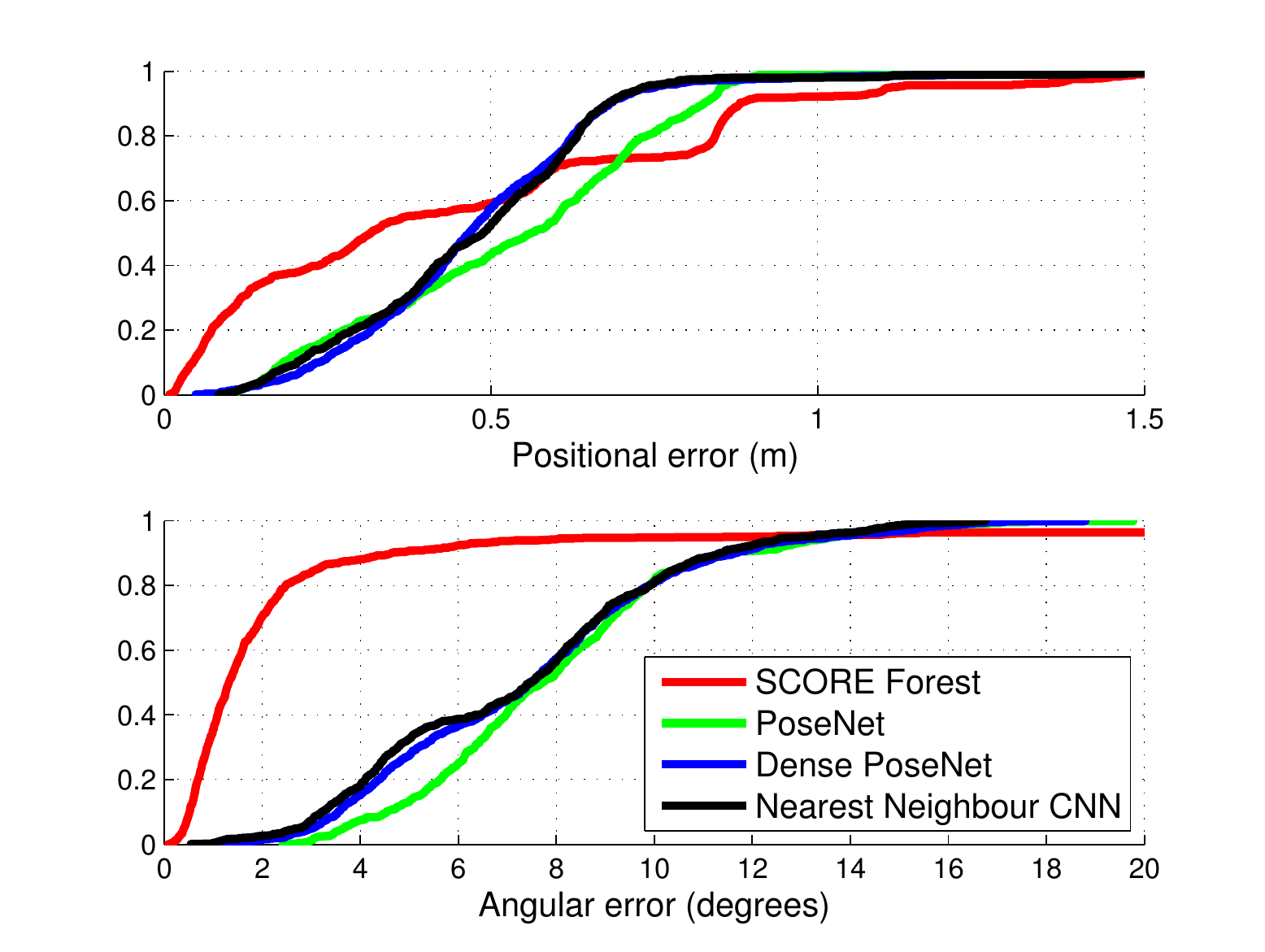}}
   	\caption{Stairs}
\end{subfigure}
   	}
   \caption{\textbf{Localization performance.} These figures show our localization accuracy for both position and orientation as a cumulative histogram of errors for the entire testing set. The regression convnet outperforms the nearest neighbour feature matching which demonstrates we regress finer resolution results than given by training. Comparing to the RGB-D SCoRe Forest approach shows that our method is competitive, but outperformed by a more expensive depth approach. Our method does perform better on the hardest few frames, above the 95th percentile, with our worst error lower than the worst error from the SCoRe approach. }
   \label{fig:histograms}
\end{figure*}

Deep learning performs extremely well on large datasets, however producing these datasets is often expensive or very labour intensive. We overcome this by leveraging structure from motion to autonomously generate training labels (camera poses). This reduces the human labour to just recording the video of each scene.

For this paper we release an outdoor urban localization dataset, \textit{Cambridge Landmarks\footnote{PoseNet code and dataset available here:\\ {\tt\small\href{http://mi.eng.cam.ac.uk/projects/relocalisation/}{mi.eng.cam.ac.uk/projects/relocalisation/}}}}, with 5 scenes. This novel dataset provides data to train and test pose regression algorithms in a large scale outdoor urban setting. A bird's eye view of the camera poses is shown in fig.~\ref{fig:map} and further details can be found in table \ref{fig:mainresults}. Significant urban clutter such as pedestrians and vehicles were present and data was collected from many different points in time representing different lighting and weather conditions. Train and test images are taken from distinct walking paths and not sampled from the same trajectory making the regression challenging (see fig.~\ref{fig:zoommap}). We release this dataset for public use and hope to add scenes to this dataset as this project progresses. 

The dataset was generated using structure from motion techniques \cite{wu2013towards} which we use as ground truth measurements for this paper. A Google LG Nexus 5 smartphone was used by a pedestrian to take high definition video around each scene. This video was subsampled in time at 2Hz to generate images to input to the SfM pipeline. There is a spacing of about 1m between each camera position.

To test on indoor scenes we use the publically available \textit{7 Scenes} dataset \cite{shotton2013scene}, with scenes shown in fig.~\ref{fig:7scenes}. This dataset contains significant variation in camera height and was designed for RGB-D relocalization. It is extremely challenging for purely visual relocalization using SIFT-like features, as it contains many ambiguous textureless features.

\section{Experiments}

\afterpage{\clearpage}
\begin{figure*}[p]
\begin{center}
\begin{subfigure}[b]{\textwidth}
\makebox[\textwidth][c]{
   	\includegraphics[width=0.16\linewidth]{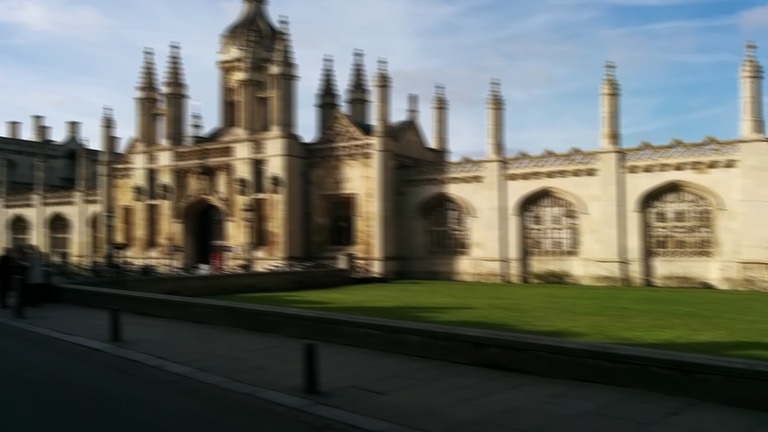}
   	\includegraphics[width=0.16\linewidth]{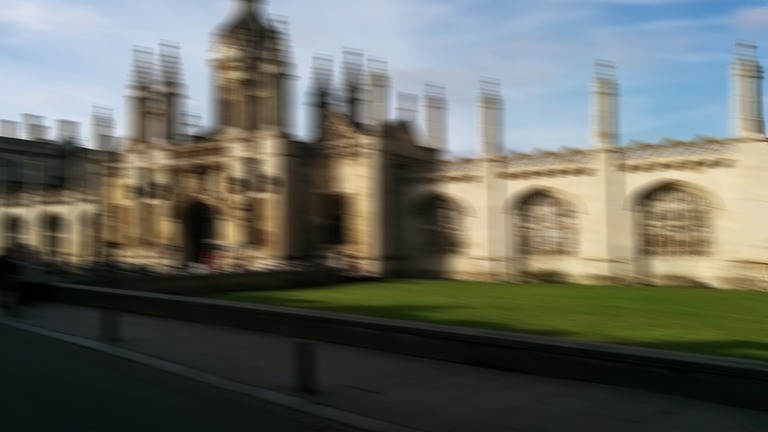}
   	\includegraphics[width=0.16\linewidth]{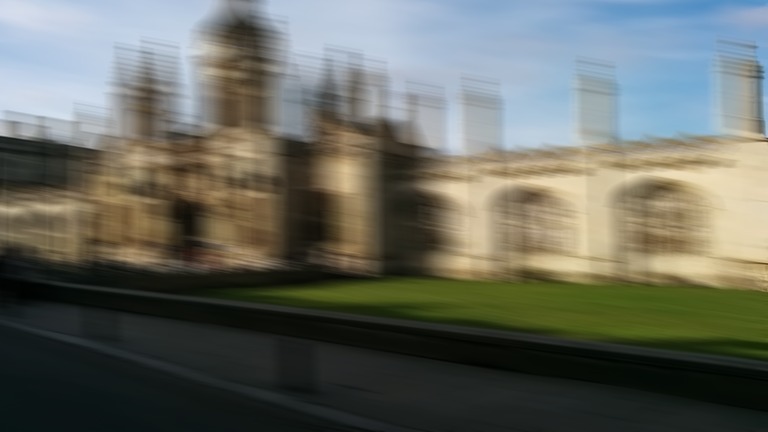}
   	\includegraphics[width=0.16\linewidth]{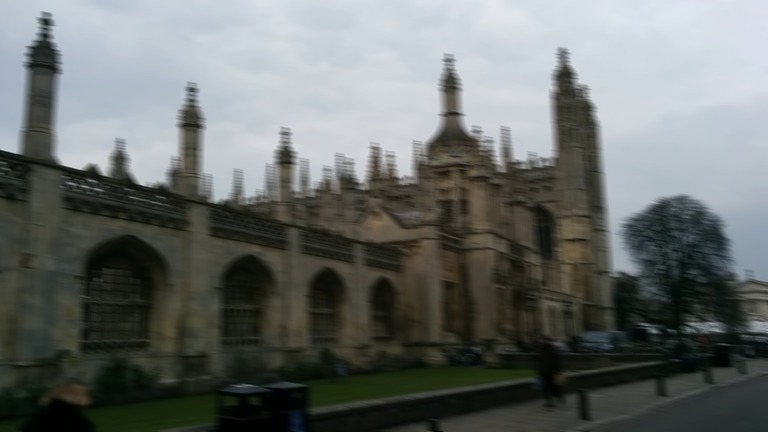}
   	\includegraphics[width=0.16\linewidth]{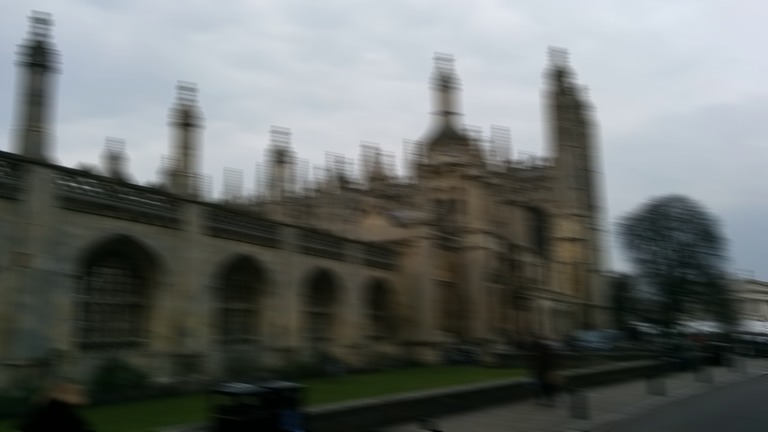}
   	\includegraphics[width=0.16\linewidth]{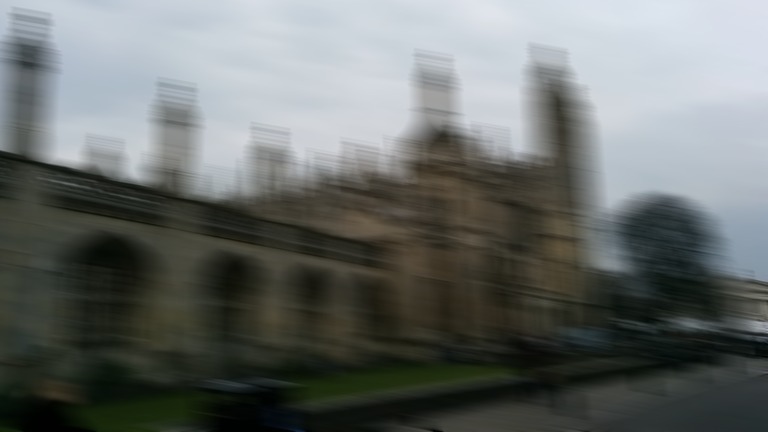}
   	}
\makebox[\textwidth][c]{
   	\includegraphics[width=0.16\linewidth]{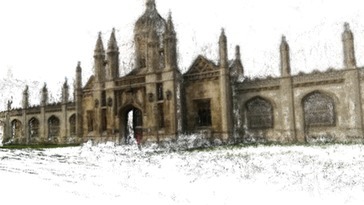}
   	\includegraphics[width=0.16\linewidth]{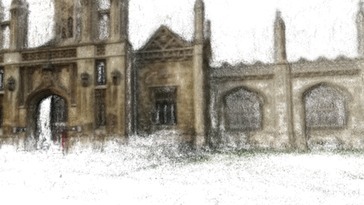}
   	\includegraphics[width=0.16\linewidth]{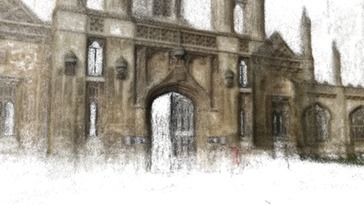}
   	\includegraphics[width=0.16\linewidth]{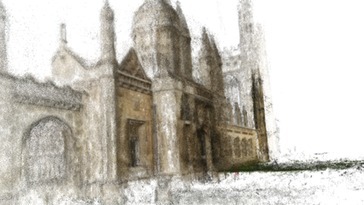}
   	\includegraphics[width=0.16\linewidth]{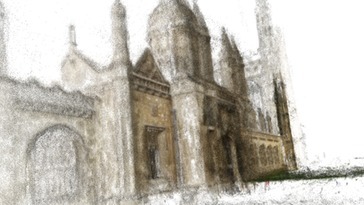}
   	\includegraphics[width=0.16\linewidth]{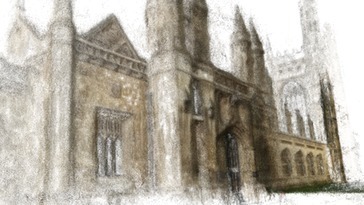}
   	}
   \caption{Relocalization with increasing levels of motion blur. The system is able to recognize the pose as high level features such as the contour outline still exist. Blurring the landmark increases apparent contour size and the system believes it is closer. }
\end{subfigure}

\begin{subfigure}[b]{\textwidth}
\makebox[\textwidth][c]{
   	\includegraphics[width=0.24\linewidth]{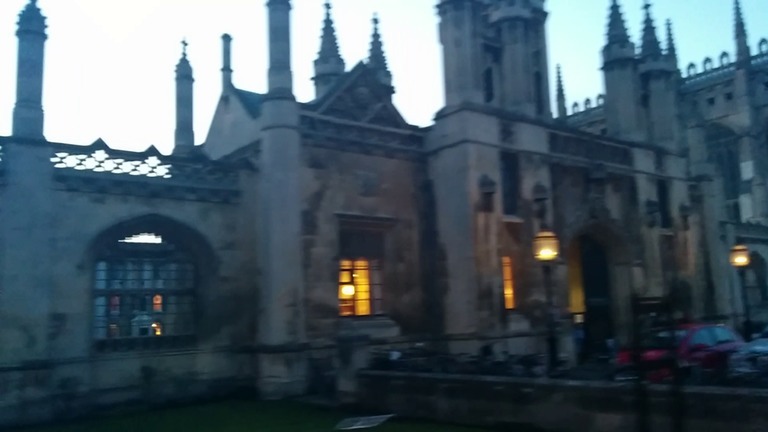}
   	\includegraphics[width=0.24\linewidth]{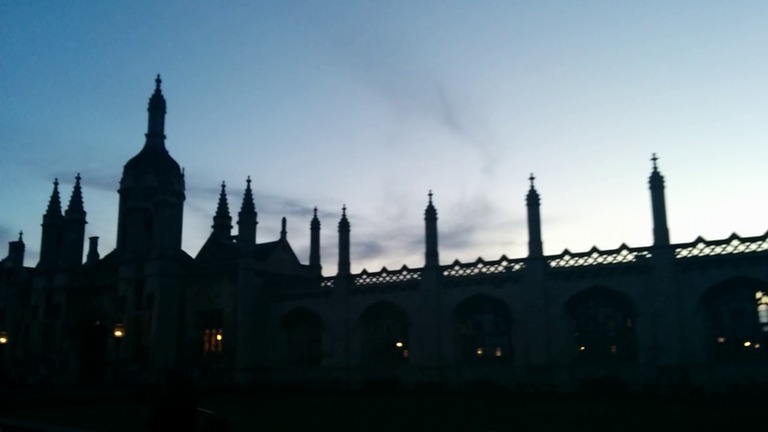}
   	\includegraphics[width=0.24\linewidth]{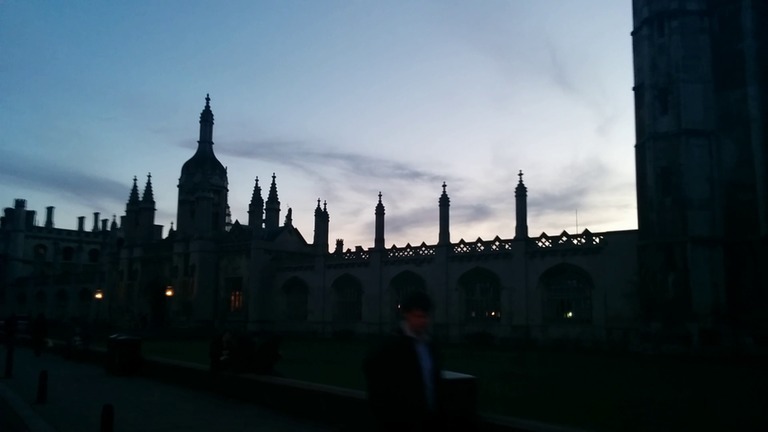}
   	\includegraphics[width=0.24\linewidth]{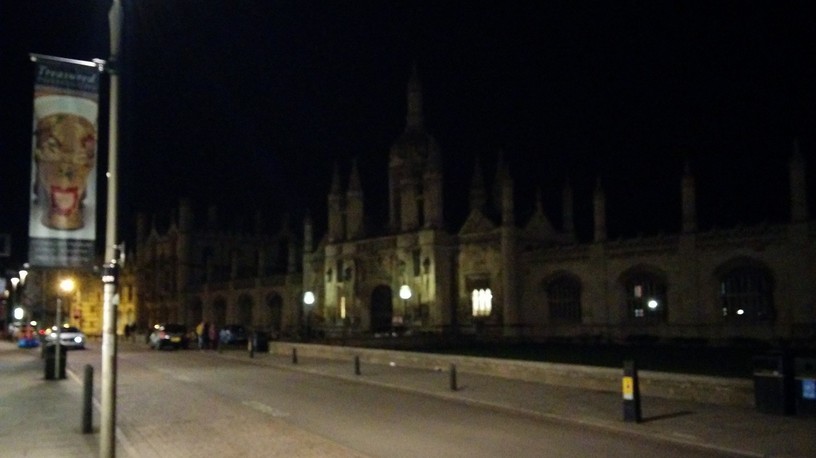}
   	}
\makebox[\textwidth][c]{
   	\includegraphics[width=0.24\linewidth]{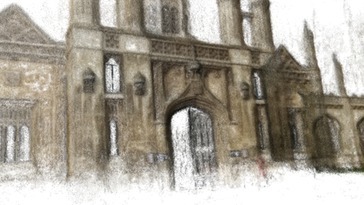}
   	\includegraphics[width=0.24\linewidth]{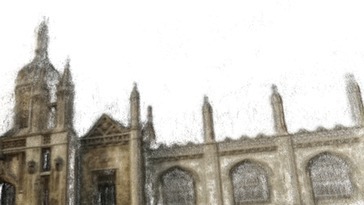}
   	\includegraphics[width=0.24\linewidth]{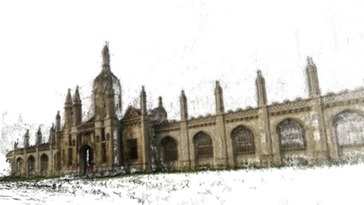}
   	\includegraphics[width=0.24\linewidth]{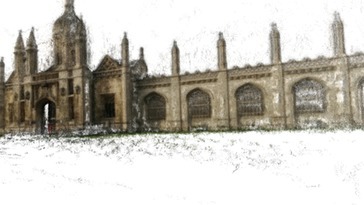}
   	}
\makebox[\textwidth][c]{
   	\includegraphics[width=0.24\linewidth]{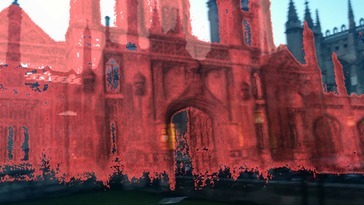}
   	\includegraphics[width=0.24\linewidth]{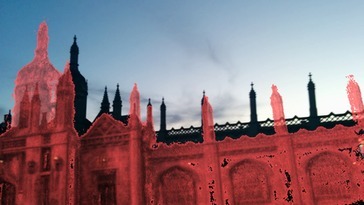}
   	\includegraphics[width=0.24\linewidth]{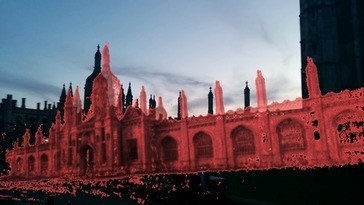}
   	\includegraphics[width=0.24\linewidth]{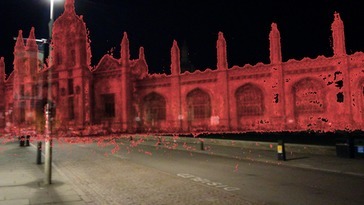}
   	}
   \caption{Relocalization under difficult dusk and night lighting conditions. In the dusk sequences, the landmark is silhouetted against the backdrop however again the convnet seems to recognize the contours and estimate pose.}
\end{subfigure}

\begin{subfigure}[t]{0.3\textwidth}
\makebox[\textwidth][c]{
   	\includegraphics[width=0.5\textwidth]{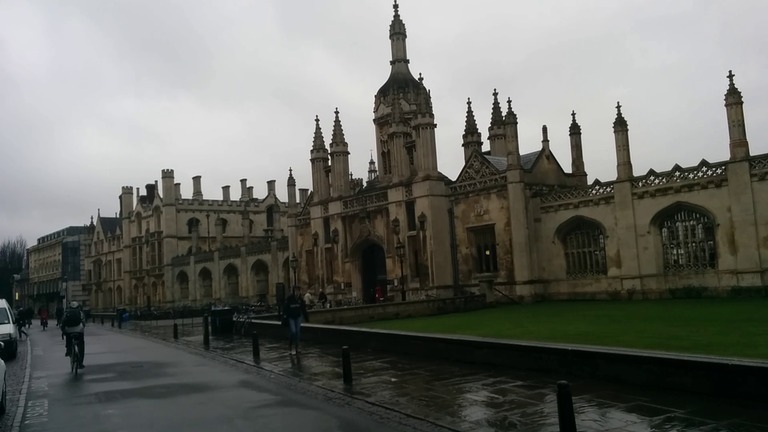}
   	\includegraphics[width=0.5\textwidth]{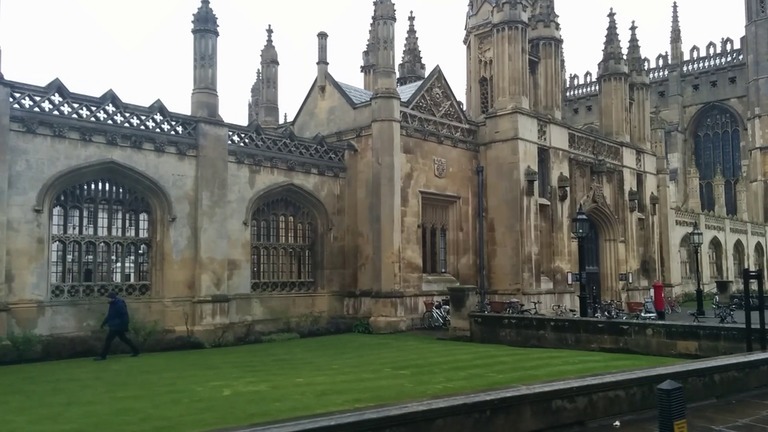}
   	}
\makebox[\textwidth][c]{
   	\includegraphics[width=0.5\textwidth]{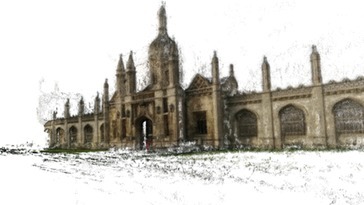}
   	\includegraphics[width=0.5\textwidth]{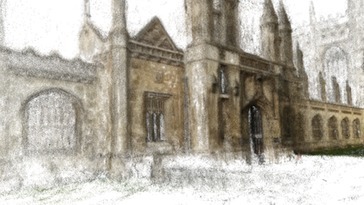}
   	}
\makebox[\textwidth][c]{
   	\includegraphics[width=0.5\textwidth]{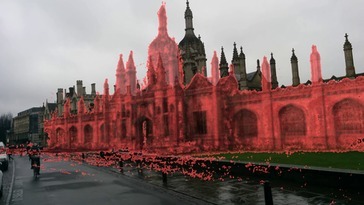}
   	\includegraphics[width=0.5\textwidth]{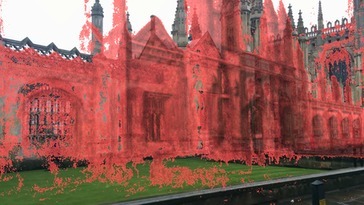}
   	}
   \caption{Relocalization with different weather conditions. PoseNet is able to effectively estimate pose in fog and rain.}
\end{subfigure}
\qquad
\begin{subfigure}[t]{0.3\textwidth}
\makebox[\textwidth][c]{
   	\includegraphics[width=0.5\textwidth]{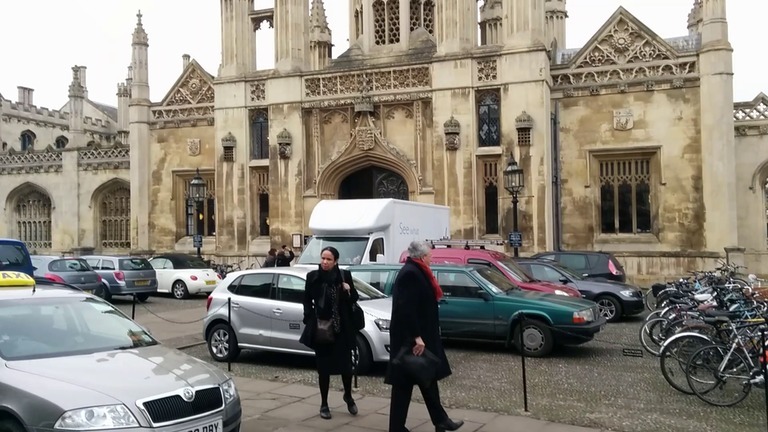}
   	\includegraphics[width=0.5\textwidth]{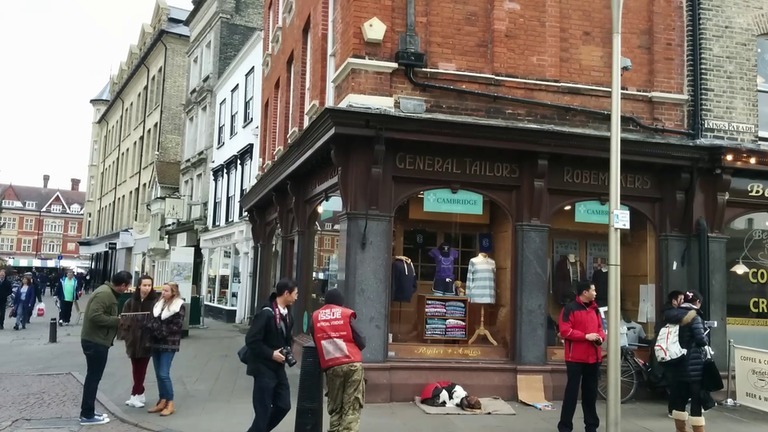}
   	}
\makebox[\textwidth][c]{
   	\includegraphics[width=0.5\textwidth]{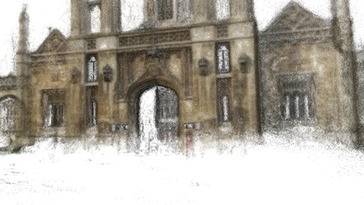}
   	\includegraphics[width=0.5\textwidth]{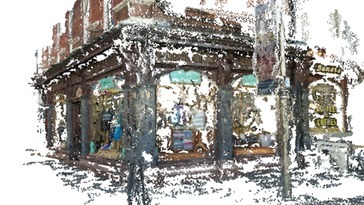}
   	}
\makebox[\textwidth][c]{
   	\includegraphics[width=0.5\textwidth]{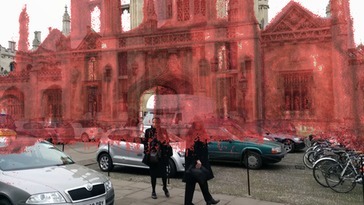}
   	\includegraphics[width=0.5\textwidth]{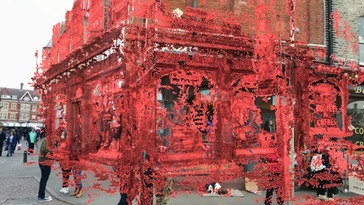}
   	}
   \caption{Relocalization with significant people, vehicles and other dynamic objects.}
\end{subfigure}
\qquad
\begin{subfigure}[t]{0.3\textwidth}
\makebox[\textwidth][c]{
   	\includegraphics[width=0.5\textwidth]{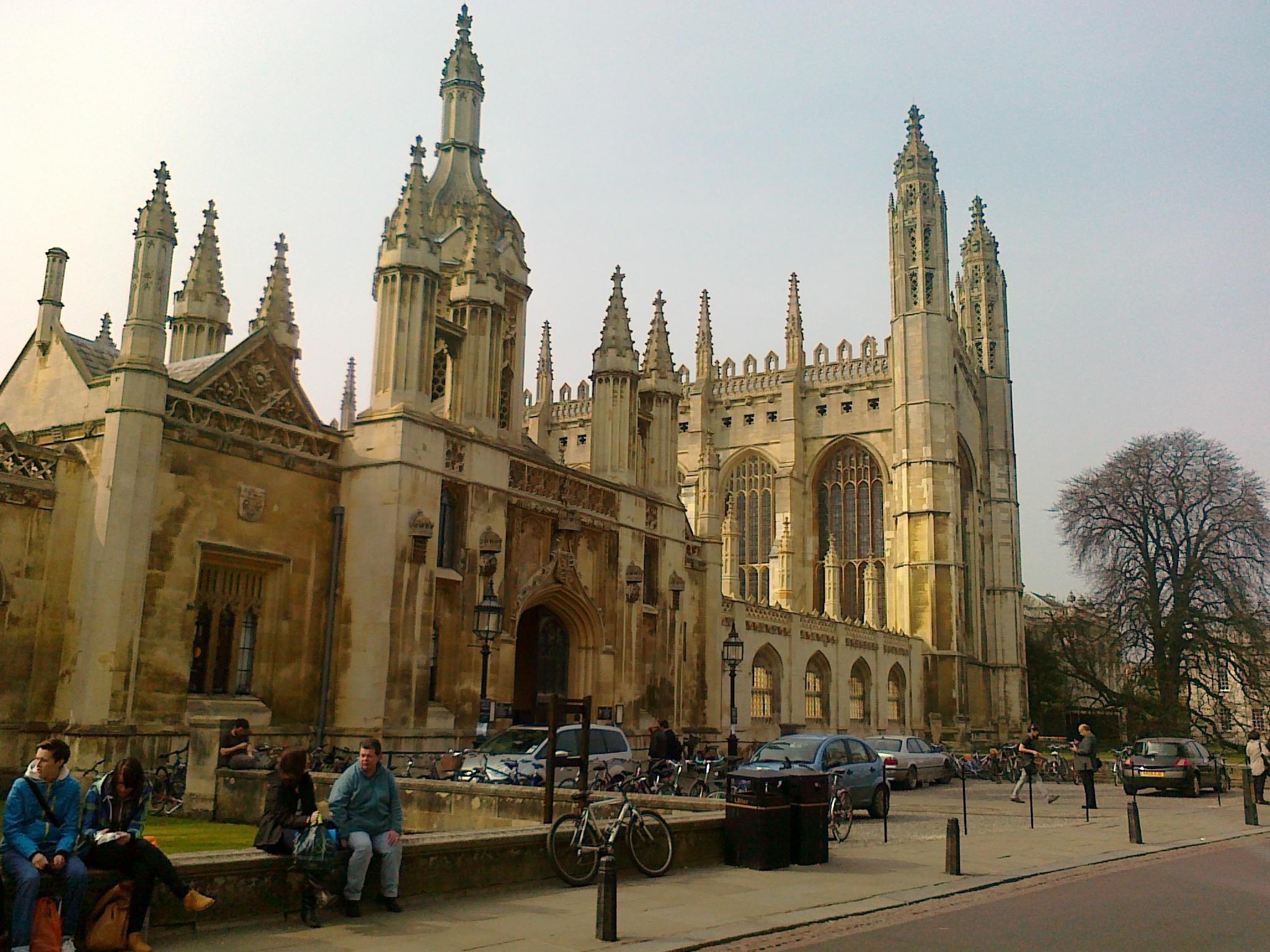}
   	\includegraphics[width=0.5\textwidth]{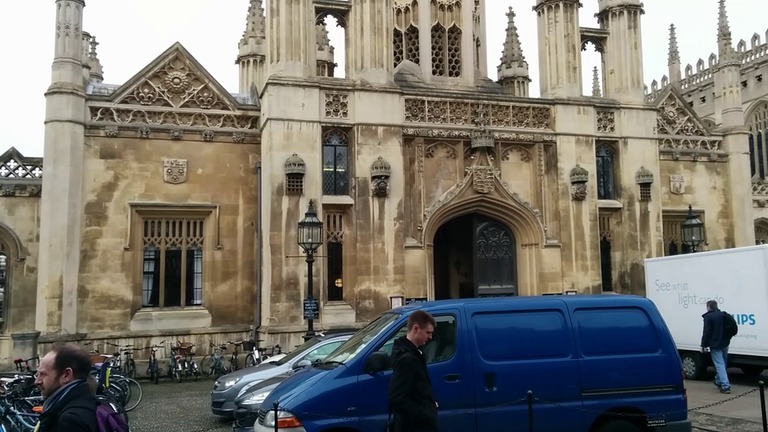}
   	}
\makebox[\textwidth][c]{
   	\includegraphics[width=0.5\textwidth]{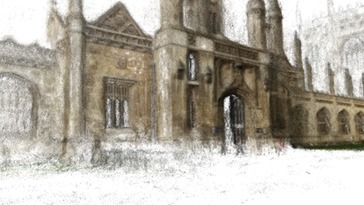}
   	\includegraphics[width=0.5\textwidth]{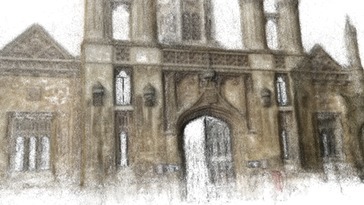}
   	}
\makebox[\textwidth][c]{
   	\includegraphics[width=0.5\textwidth]{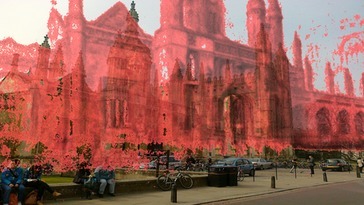}
   	\includegraphics[width=0.5\textwidth]{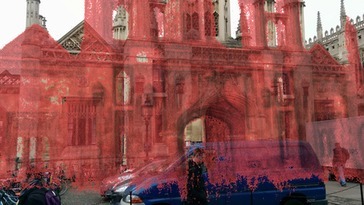}
   	}
   \caption{Relocalization with unknown camera intrinsics: SLR with focal length 45mm (left), and iPhone 4S with focal length 35mm (right) compared to the dataset's camera which had a focal length of 30mm.}
\end{subfigure}
\end{center}
	\caption{\textbf{Robustness to challenging real life situations.} Registration with point based techniques such as SIFT fails in examples (a-c), therefore ground truth measurements are not available. None of these types of challenges were seen during training. As convnets are able to understand objects and contours they are still successful at estimating pose from the building's contour in the silhouetted examples (b) or even under extreme motion blur (a). Many of these quasi invariances were enhanced by pretraining from the scenes dataset.}
	\label{fig:difficultexamples}
\end{figure*}

We show that PoseNet is able to effectively localize across both the indoor \textit{7 Scenes} dataset and outdoor \textit{Cambridge Landmarks} dataset in table \ref{fig:mainresults}. To validate that the convnet is regressing pose beyond that of the training examples we show the performance for finding the nearest neighbour representation in the training data from the feature vector produced by the localization convnet. As our performance exceeds this we conclude that the convnet is successfully able to regress pose beyond training examples (see fig.~\ref{fig:zoommap}). We also compare our algorithm to the RGB-D SCoRe Forest algorithm \cite{shotton2013scene}. 

Fig.~\ref{fig:histograms} shows cumulative histograms of localization error for two indoor and two outdoor scenes. We note that although the SCoRe forest is generally more accurate, it requires depth information, and uses higher-resolution imagery. The indoor dataset contains many ambiguous and textureless features which make relocalization without this depth modality extremely difficult. We note our method often localizes the most difficult testing frames, above the 95th percentile, more accurately than SCoRe across all the scenes. We also observe that dense cropping only gives a modest improvement in performance. It is most important in scenes with significant clutter like pedestrians and cars, for example King's College, Shop Fa\c cade and St Mary's Church.

We explored the robustness of this method beyond what was tested in the dataset with additional images from dusk, rain, fog, night and with motion blur and different cameras with unknown intrinsics. Fig.~\ref{fig:difficultexamples} shows the convnet generally handles these challenges well. SfM with SIFT fails in all these cases so we were not able to generate a ground truth camera pose, however we infer the accuracy by viewing the 3D reconstruction from the predicted camera pose, and overlaying this onto the input image.

\subsection{Robustness against training image spacing}

\begin{figure}[t]
\begin{center}
   	\includegraphics[width=\linewidth]{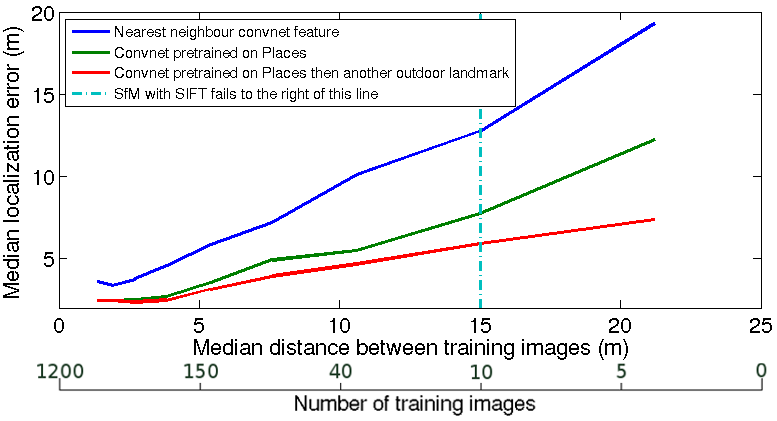}
\end{center}
   \caption{\textbf{Robustness to a decreasing training baseline} for the King's College scene. Our system exhibits graceful decline in performance as fewer training samples are used.}
\label{fig:baseline}
\end{figure}

We demonstrate in fig.~\ref{fig:baseline} that, for an outdoor scale scene, we gain little by spacing the training images more closely than 4m. The system is robust to very large spatial separation between training images, achieving reasonable performance even with only a few dozen training samples. The pose accuracy deteriorates gracefully with increased training image spacing, whereas SIFT-based SfM sharply fails after a certain threshold as it requires a small baseline \cite{lowe2004distinctive}.

\subsection{Importance of transfer learning}

\begin{figure}[t]
\begin{center}
   	\adjustbox{trim={0} {0} {0} {0.05\height},clip}{\includegraphics[width=0.9\linewidth]{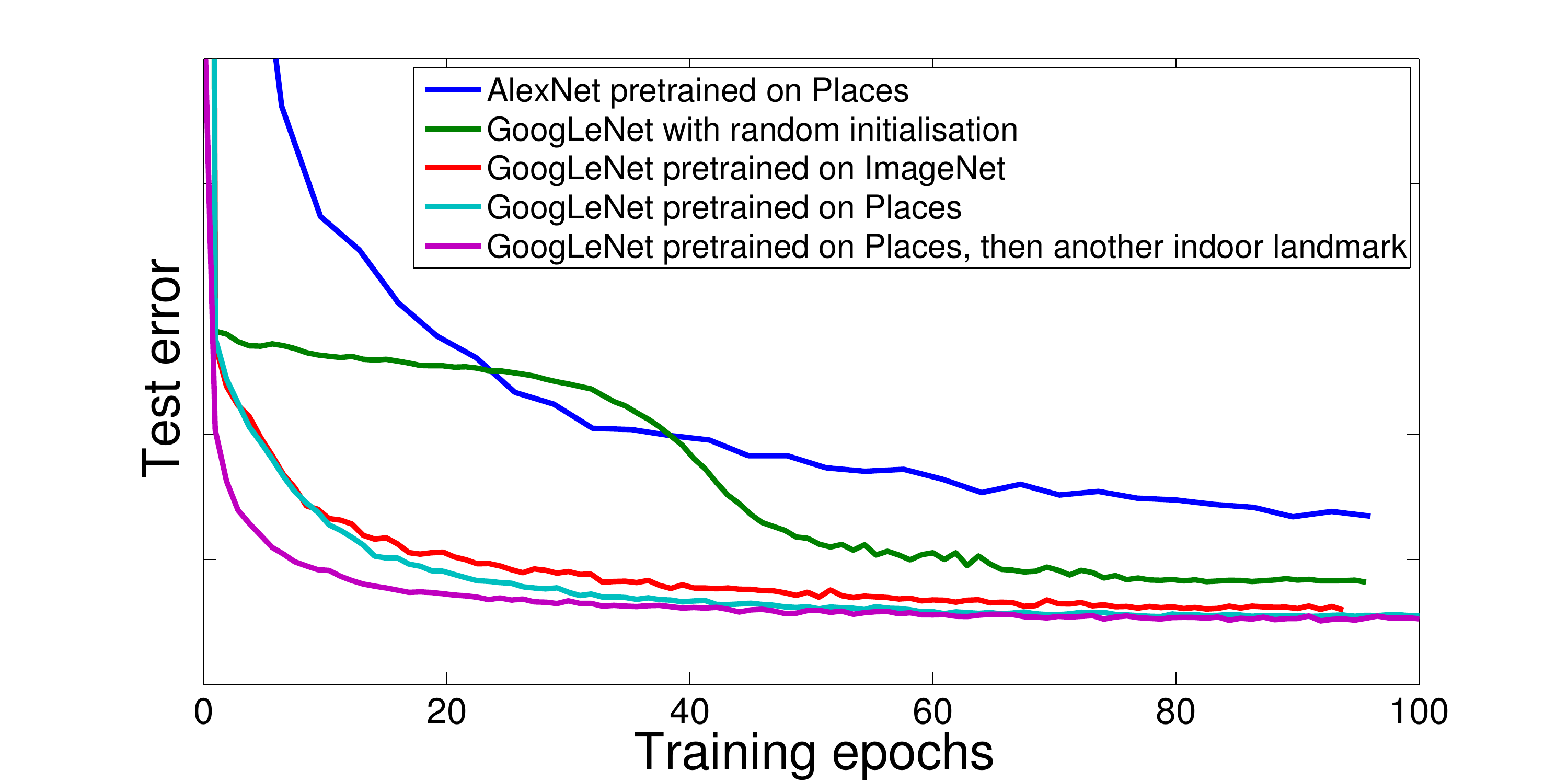}}
\end{center}
   \caption{\textbf{Importance of transfer learning.} Shows how pretraining on large datasets gives an increase in both performance and training speed.}
\label{fig:transfer}
\end{figure}

In general convnets require large amounts of training data. We sidestep this problem by starting our pose training from a network pretrained on giant datasets such as \textit{ImageNet} and \textit{Places}. Similar to what has been demonstrated for classification tasks, fig.~\ref{fig:transfer} shows how transfer learning can be utilised effectively between classification and complicated regression tasks. Such `transfer learning' has been demonstrated elsewhere for training classifiers \cite{razavian2014cnn,oquab2014learning,bengio2013representation}, but here we demonstrate transfer learning from classification to the qualitatively different task of pose regression. It is not immediately obvious that a network trained to output pose-invariant classification labels would be suitable as a starting point for a pose regressor. We find, however, that this is not a problem in practice. A possible explanation is that, in order for its output to be invariant to pose, the classifier network must keep track of pose, to better factor its effects away from identity cues. This would agree with our own findings that a network trained to output position and orientation outperforms a network trained to output only position. By preserving orientation information in the intermediate representations, it is better able to factor the effects of orientation out of the final position estimation. Transfer learning gives not only a large improvement in training speed, but also end performance. 

The relevance of data is also important. In fig.~\ref{fig:transfer} the \textit{Places} and \textit{ImageNet} curves initially have the same performance. However, ultimately the \textit{Places} pretraining performs better due to being a more relevant dataset to this localization task.

\subsection{Visualising features relevant to pose}

\begin{figure*}[t]
\makebox[\linewidth][c]{
   	\includegraphics[width=0.33\linewidth]{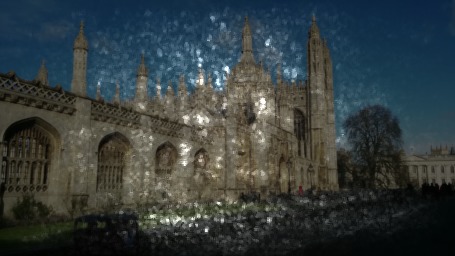}
   	\includegraphics[width=0.33\linewidth]{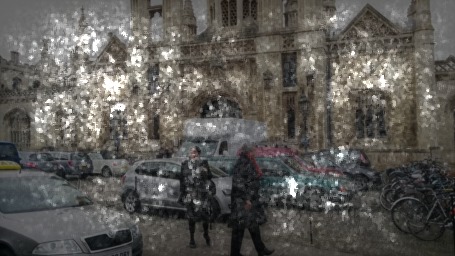}
   	\includegraphics[width=0.33\linewidth]{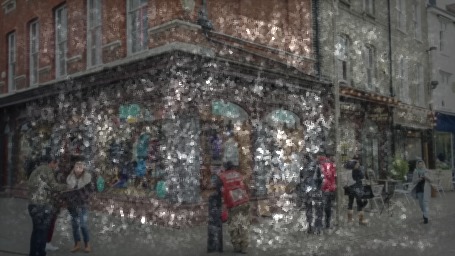}
   	}
   \caption{\textbf{Saliency maps.} This figure shows the saliency map superimposed on the input image. The saliency maps suggest that the convnet exploits not only distinctive point features (\`{a} la SIFT), but also large textureless patches, which can be as informative, if not more so, to the pose. This, combined with a tendency to disregard dynamic objects such as pedestrians, enables it to perform well under challenging circumstances. (Best viewed electronically.)}
\label{fig:saliency}
\end{figure*}

Fig.~\ref{fig:saliency} shows example saliency maps produced by PoseNet. The saliency map, as used in \cite{simonyan2013deep}, is the magnitude of the gradient of the loss function with respect to the pixel intensities. This uses the sensitivity of the pose with respect to the pixels as an indicator of how important the convnet considers different parts of the image.

These results show that the strongest response is observed from higher-level features such as windows and spires. However a more surprising result is that PoseNet is also very sensitive to large textureless patches such as road, grass and sky. These textureless patches may be more informative than the highest responding points because the effect of a group of pixels on the pose variable is the sum of the saliency map values over that group of pixels. This evidence points to the net being able to localize off information from these textureless surfaces, something which interest-point based features such as SIFT or SURF fail to do.

The last observation is that PoseNet has an attenuated response to people and other noisy objects, effectively masking them. These objects are dynamic, and the convnet has identified them as not appropriate for localization.

\subsection{Viewing the internal representation}

\begin{figure}[t]
\makebox[\linewidth][c]{
   	\framebox[\width]{ \includegraphics[width=0.35\linewidth]{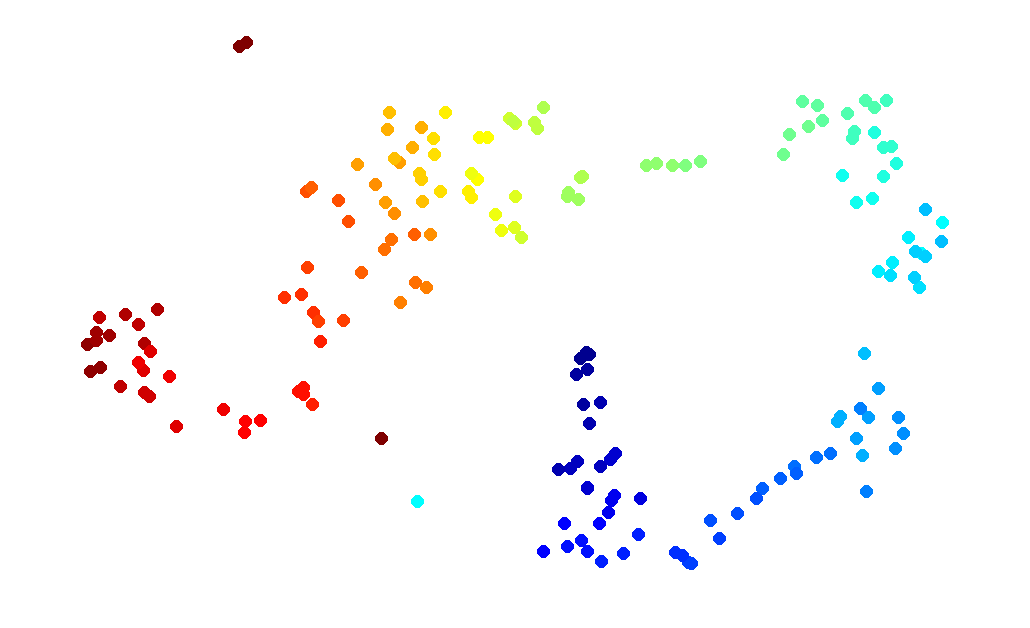}}
   	\framebox[\width]{ \includegraphics[width=0.35\linewidth]{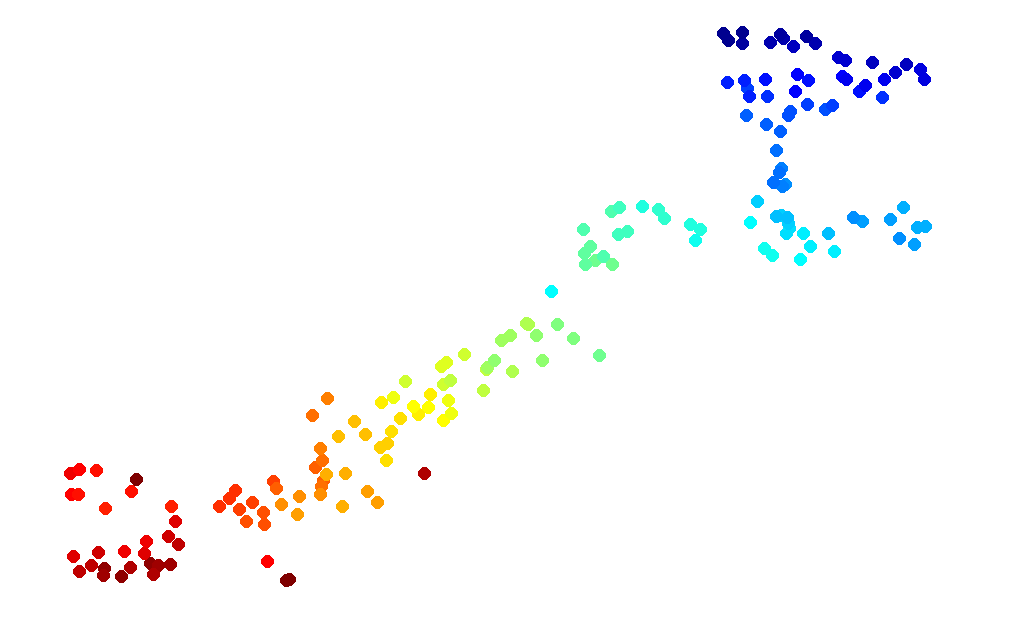}}
   	\framebox[\width]{ \includegraphics[width=0.35\linewidth]{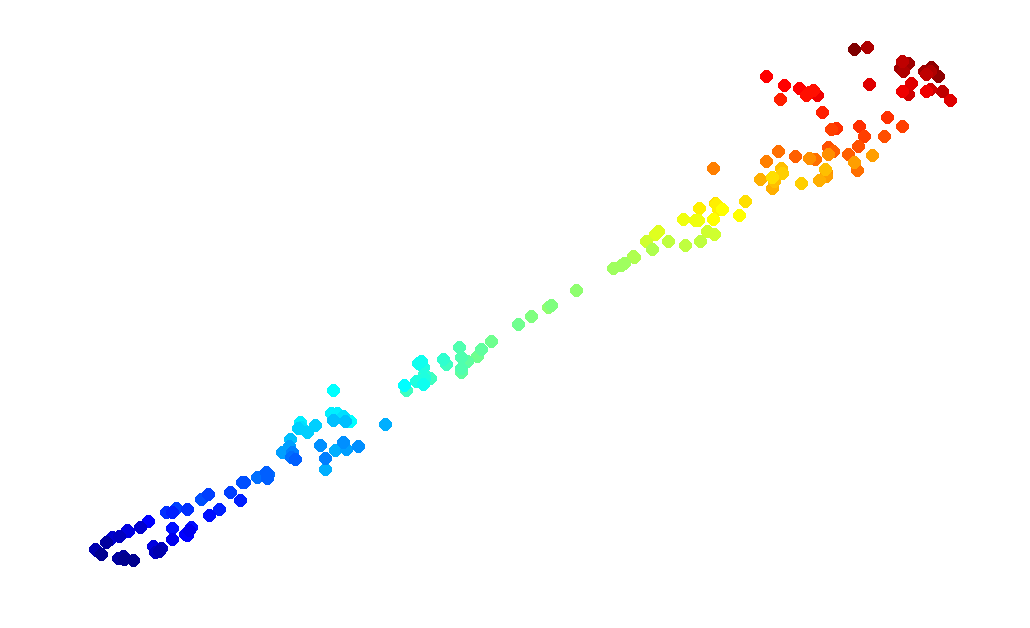}}
   	}
\makebox[\linewidth][c]{
\makebox[0.35\linewidth][c]{
(a)
   	}
\makebox[0.35\linewidth][c]{
(b)
   	}
\makebox[0.35\linewidth][c]{
(c)
   	}}
   \caption{\textbf{Feature vector visualisation.} t-SNE visualisation of the feature vectors from a video sequence traversing an outdoor scene (King's College) in a straight line. Colour represents time. The feature representations are generated from the convnet with weights trained on \textit{Places} (a), \textit{Places} then another outdoor scene, St Mary's Church (b), \textit{Places} then this outdoor scene, King's College (c). Despite (a,b) not being trained on this scene, these visualizations suggest that it is possible to compute the pose as a simple, if non-linear, function of these representations.}
\label{fig:tsne}
\end{figure}

t-SNE \cite{van2008visualizing} is an algorithm for embedding high-dimensional data in a low dimensional space, in a way that tries to preserve Euclidean distances. It is often used, as we do here, to visualize high-dimensional feature vectors in two dimensions. In fig.~\ref{fig:tsne} we apply t-SNE to the feature vectors computed from a sequence of video frames taken by a pedestrian. As these figures show, the feature vectors are a function that smoothly varies with, and is largely one-to-one with, pose. This `pose manifold' can be observed not only on networks trained on other scenes, but also networks trained on classification image sets without pose labels. This further suggests that classification convnets preserve pose information up to the final layer, regardless of whether it's expressed in the output. However, the mapping from feature vector to pose becomes more complicated for networks not trained on pose data. Furthermore, as this manifold exists on scenes that the convnet was not trained on, the convnet must learn some generic representation of the relationship between landmarks, geometry and camera motion. This demonstrates that the feature vector that is produced from regression is able to generalize to other tasks in the same way as classification convnets.

\subsection{System efficiency}

\begin{figure}[t]
\makebox[\linewidth][c]{
   	\includegraphics[width=\linewidth]{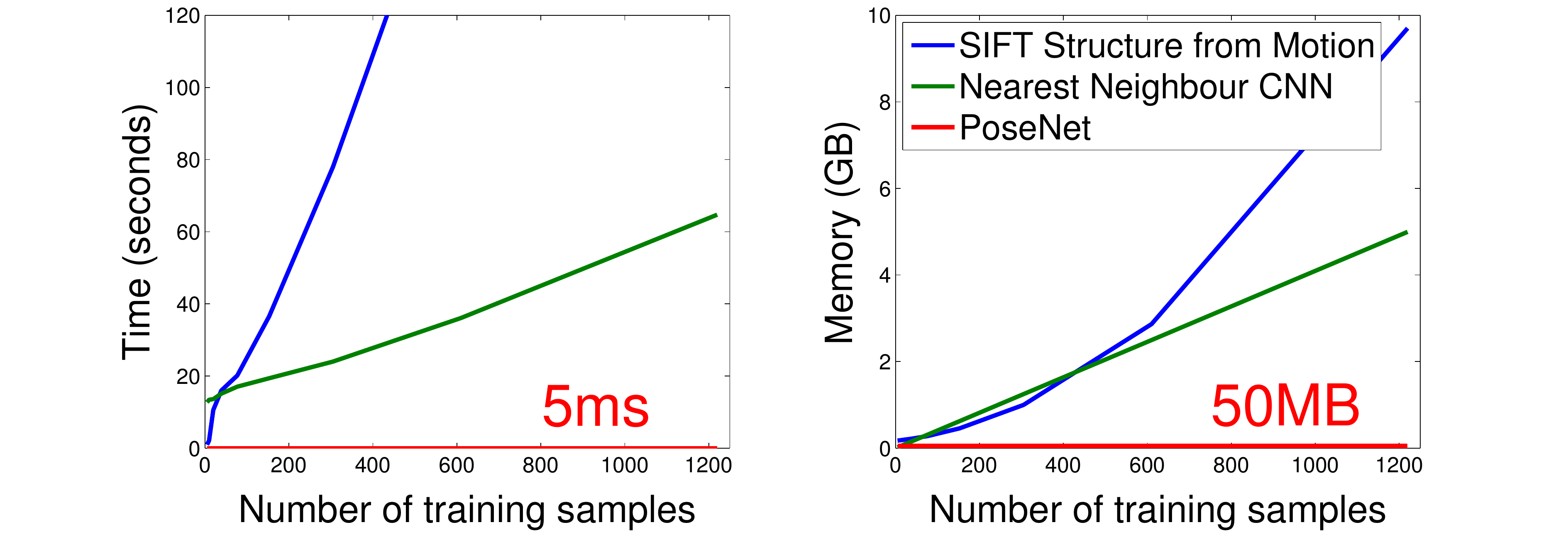}
   	}
   \caption{\textbf{Implementation efficiency.} Experimental speed and memory use of the convnet regression, nearest neighbour convnet feature vector and SIFT relocalization methods.}
\label{fig:speed}
\end{figure}

Fig.~\ref{fig:speed} compares system performance of PoseNet on a modern desktop computer. Our network is very scalable, as it only takes $50$ MB to store the weights, and $5ms$ to compute each pose, compared to the gigabytes and minutes for metric localization with SIFT. These values are independent of the number of training samples in the system while metric localization scales $\mathcal{O}(n^2)$ with training data size \cite{wu2013towards}. For comparison matching to the convnet nearest neighbour is also shown. This requires storing feature vectors for each training frame, then perform a linear search to find the nearest neighbour for a given test frame.

\section{Conclusions}

We present, to our knowledge, the first application of deep convolutional neural networks to end-to-end 6-DOF camera pose localization. We have demonstrated that one can sidestep the need for millions of training images by use of transfer learning from networks trained as classifiers. We showed that such networks preserve ample pose information in their feature vectors, despite being trained to produce pose-invariant outputs. Our method tolerates large baselines that cause SIFT-based localizers to fail sharply.

In future work, we aim to pursue further uses of multiview geometry as a source of training data for deep pose regressors, and explore probabilistic extensions to this algorithm \cite{kendall2015modelling}. It is obvious that a finite neural network has an upper bound on the physical area that it can learn to localize within. We leave finding this limit to future work.

{\small
\bibliographystyle{ieee}
\bibliography{paper}
}

\end{document}